\documentclass{article}

\usepackage[final]{fl_neurips_2022}

\usepackage[utf8]{inputenc} 
\usepackage[T1]{fontenc}    
\usepackage{hyperref}       
\usepackage{url}            
\usepackage{booktabs}       
\usepackage{amsfonts}       
\usepackage{nicefrac}       
\usepackage{microtype}      
\usepackage{xcolor}         
\usepackage{amsmath,amssymb} 

\DeclareMathOperator*{\argmin}{arg\,min}
\usepackage{xcolor}
\usepackage{siunitx}
\usepackage{booktabs} 
\usepackage{subcaption}
\usepackage{multirow}
\usepackage{adjustbox} 
\usepackage{subcaption}
\usepackage{dsfont}
\newtheorem{definition}{Definition}

\usepackage{xspace}
\newcommand{\irm}{\textsc{IRM}\xspace}
\newcommand{\irmgames}{\textsc{IRM Games}\xspace}
\newcommand{\firm}{\textsc{F-IRM Games}\xspace}
\newcommand{\virm}{\textsc{V-IRM Games}\xspace}
\newcommand{\ffl}{\textsc{F-FL Games}\xspace}
\newcommand{\vfl}{\textsc{V-FL Games}\xspace}
\newcommand{\brd}{\textsc{BRD}\xspace}
\newcommand{\fedsgd}{\textsc{FedSGD}\xspace}
\newcommand{\fedbn}{\textsc{FedBN}\xspace}
\newcommand{\fedprox}{\textsc{FedProx}\xspace}
\newcommand{\fedavg}{\textsc{FedAVG}\xspace}
\newcommand{\ffictplay}{\textsc{F-FL Games (Smooth)}\xspace}
\newcommand{\vfictplay}{\textsc{V-FL Games (Smooth)}\xspace}
\newcommand{\finalalgo}{\textsc{V-FL Games (Smooth \!\!+ \!\!Fast)}\xspace}
\newcommand{\multilinecomment}[1]{}

\newcommand{\sharutalgo}{\textsc{Algorithm}\xspace}
\newcommand{\trainacc}{\textsc{Train Accuracy}\xspace}
\newcommand{\testacc}{\textsc{Test Accuracy}\xspace}
\newcommand{\flgames}{\textsc{FL Games}\xspace}
\newcommand{\coloredmnist}{\textsc{Colored MNIST}\xspace}
\newcommand{\extendedcoloredmnist}{\textsc{Extended Colored MNIST}\xspace}
\newcommand{\coloredfashionmnist}{\textsc{Colored Fashion MNIST}\xspace}

\newcommand{\cifar}{\textsc{Spurious CIFAR10}\xspace}
\newcommand{\dsprites}{\textsc{Colored Dsprites}\xspace}

\usepackage{algorithm}
\usepackage[noend]{algpseudocode}
\algdef{SE}[SUBALG]{Indent}{EndIndent}{}{\algorithmicend\ }%
\algtext*{Indent}
\algtext*{EndIndent}

\newcommand{\RNum}[1]{\uppercase\expandafter{\romannumeral #1\relax}}
\usepackage{natbib}

\title{FL Games: A Federated Learning Framework for Distribution Shifts}

\author{%
  Sharut Gupta\\
  Mila, Universite de Montreal\\
  Imagia Cybernetics Inc. \\
  Indian Institute of Technology Delhi\\
  \texttt{sharut.gupta@mila.quebec} \\
   \And
   Kartik Ahuja \\
   Mila, Universite de Montreal\\
   \texttt{kartik.ahuja@mila.quebec} \\
   \AND
   Mohammad Havaei  \\
   Imagia Cybernetics Inc. \\
   \texttt{mohammad@imagia.com} \\
   \And
   Niladri Chatterjee  \\
   Indian Institute of Technology Delhi\\
   \texttt{ niladri@maths.iitd.ac.in} \\
   \And
   Yoshua Bengio \\
   Mila, Universite de Montreal\\
   \texttt{yoshua.bengio@mila.quebec} \\
}

\begin{document}

\maketitle

\begin{abstract}
   Federated learning aims to train predictive models for data that is distributed across clients, under the orchestration of a server. 
However, participating clients typically each hold data from a different distribution, which can yield to catastrophic generalization on data from a different client, which represents a new domain.  In this work, we argue that in order to generalize better across non-i.i.d. clients, it is imperative to only learn correlations that are stable and invariant across domains. We propose \flgames, a game-theoretic framework for federated learning that learns causal features that are invariant across clients. While training to achieve the Nash equilibrium, the traditional best response strategy suffers from high-frequency oscillations. We demonstrate that \flgames effectively resolves this challenge and exhibits smooth performance curves. Further, \flgames scales well in the number of clients, requires significantly fewer communication rounds, and is agnostic to device heterogeneity. Through empirical evaluation, we demonstrate that \flgames achieves high out-of-distribution performance on various benchmarks.\\
\end{abstract}


\section{Introduction}
With the rapid advance in technology and growing prevalence of smart devices, Federated Learning (FL) has emerged as an attractive distributed learning paradigm for machine learning models over networks of computers~\cite{kairouz2019advances,li2020federated,bonawitz2019towards}. In FL, multiple sites with local data, often known as \emph{clients}, collaborate to jointly train a shared model under the orchestration of a central hub called the \emph{server} while keeping their data private.  

\par While FL serves as an attractive alternative to centralized training because the client data does not need to move to the server, \emph{statistical heterogeneity} is a key challenge in its optimization.
While one of the most popular algorithms in this setup, Federated Averaging (\fedavg)~\cite{mcmahan2017communication}  delivers huge communication gains in i.i.d. (independent and identically distributed) setting, its performance on non-i.i.d. clients is an active area of research. As shown by \cite{karimireddy2020scaffold}, client heterogeneity has direct implications on the convergence of \fedavg since it introduces a \emph{drift} in the updates of each client with respect to the server model. 
While recent works \cite{li2019feddane,karimireddy2020scaffold,yu2019linear,wang2020tackling,li2020federated,lin2020ensemble,li2019fedmd,zhu2021data} have tried to address client heterogeneity through constrained gradient optimization and knowledge distillation, most did not tackle the underlying distribution shift. 
These methods can at best generalize to interpolated domains and fail to extrapolate well, i.e., generalize to new extrapolated domains~\footnote{Similar to \cite{krueger2021out}, we define interpolated domains as the domains which fall within the convex hull of training domains and extrapolated domains as those that fall outside of that convex hull.}. 

\par 
Over the past year, there has been a surge in interest in bringing the machinery of causality into machine learning \cite{arjovsky2019invariant,ahuja2020invariant,scholkopf2019causality,ahuja2021linear,parascandolo2020learning,robey2021model,krueger2021out,rahimian2019distributionally}.
However, despite their success, they suffer from key limitations which are unsuitable for deployment in a real-world setup. 
\par Since FL typically consists of a large number of clients, it is natural for data at each client to represent different annotation tools, measuring circumstances, experimental environments, and external interventions.
Inspired by this idea and by the recent progress in causal machine learning, we draw connections between OOD generalization and robustness across heterogeneous clients in FL. 
To date, only two scientific works \cite{francis2021towards,tenison2021gradient} have incorporated the learning of invariant predictors in order to achieve strong generalization in FL.
The former adapts masked gradients \cite{parascandolo2020learning} and the latter builds on \irm to exploit invariance and improve leakage protection in FL. While \irm lacks theoretical convergence guarantees, failure modes of \cite{parascandolo2020learning} like formation of dead zones and high sensitivity to small perturbations \cite{shahtalebi2021sand} are also issues when it is applied in FL, rendering it unreliable.
\par In this study, we consider \irmgames \cite{ahuja2020invariant} from the OOD generalization literature since its formulation shows resemblance to the standard FL setup. However, as discussed above, \irmgames too encounters a few fundamental challenges not just specific to FL but also in a generic ML framework. We take a step towards fixing these limitations and addressing the challenge of client heterogeneity under distribution shifts in FL from a causal viewpoint. Specifically, we propose Federated Learning Games (\flgames) for learning causal representations which are stable across clients. We summarize the our main contributions below.

\begin{itemize}
    \item We propose a new framework called \flgames for learning causal representations that are invariant  across clients in a federated learning setup.
    \item The underlying sequential game theoretic framework in \irmgames causes the time complexity of FL algorithm to scale linearly with the number of clients. Inspired from the game theory literature, we equip our algorithm to allow parallel updates across clients, further resulting in superior scalability.
    \item \irmgames exhibits large oscillations in the performance metrics as the training progresses, making it difficult to define a valid stopping criterion. Using ensembles over client's historical actions, we demonstrate that \flgames appreciably smoothens these oscillations.
    \item The convergence rate of \irmgames is slow and hence directly impacts systems with speed or communication cost as a primary bottleneck. By increasing the local computation at each client, we show that \flgames exhibits high communication efficiency
    \item Empirically, we show that the performance of the invariant predictors found by our approach on unseen OOD clients improves significantly over state-of-art prior works.
    
\end{itemize}

\section{Background}

\subsection{Causality in Machine Learning} \label{background_causality}
Consider a multi-source domain generalization task, where the goal is to learn a robust set of parameters that generalize well to unseen (test) environments $\mathcal{E}_{all} \supset \mathcal{E}_{tr}$, given a set of $m$ training domains (or environments) $\mathcal{E}_{tr}$. A popular algorithm, Invariant Risk Minimization (\irm) \cite{arjovsky2019invariant}
does so by defining the Invariance principle
\begin{definition} A data representation $\phi: \mathcal{X} \rightarrow \mathcal{Z}$ elicits an invariant predictor $w \circ \phi$ across environments $\mathcal{E}_{tr}$ if there is a classifier $w: \mathcal{Z} \rightarrow \mathcal{Y}$ simultaneously optimal for all environments i.e. $w \in \argmin_{w'\in \mathcal{Z}\rightarrow \mathcal{Y}}R^e(w' \circ \phi), \forall e\in \mathcal{E}_{tr}$ 
\end{definition}


\par Invariant Risk Minimization Games (\irmgames) propose a game theoretic formulation for finding $(\phi,w)$ that satisfy the invariance principle. It endows each environment with its own predictor $w^k \in \mathcal{H}_w$ and aims to train an ensemble model $w^{av}(z) = \frac{1}{|\mathcal{E}_{tr}|}\sum_{k=1}^{|\mathcal{E}_{tr}|} w^k(z)$ for each $z \in \mathcal{Z}$ s.t. $w^{av}$ satisfies the following optimization problem
\begin{equation} \label{eq:irmgames_bilevel}
\min_{w^{av},\phi\in \mathcal{H}_{\phi}} \sum_k R^k(w^{av} \circ \phi)
 \text {s.t. } w^k \in \argmin_{w_k' \in \mathcal{H}_w}  R^k\Bigg( \frac{1}{|\mathcal{E}_{tr}|}(w_k' + \sum_{\substack{q \in \mathcal{E}_{tr} \\ q \neq k}} w^q)\circ \phi\Bigg), \forall k \in \mathcal{E}_{tr}
\end{equation}
where $\mathcal{H}_{\phi}$ , $\mathcal{H}_w$ are the hypothesis sets for feature extractors and predictors, respectively.
The constraint in equation \ref{eq:irmgames_bilevel} is equivalent to the Nash equilibrium of a game with each environment $k$ as a player with action $w^k$, playing to maximize its utility $-R^k(w^{av}, \phi)$. The resultant game is solved using the best response dynamics (\brd) with clockwise updates (for more details, refer to the supplement) and is referred to as \virm. Fixing $\phi$ to an identity map in \virm is also shown to be very effective and is called \firm. 

\section{Federated Learning Games (\flgames)}
OOD generalization is often typified using the notion of data-generating environments. \cite{arjovsky2019invariant} formalizes an environment as a data generating distribution representing a particular location, time, context, circumstances and so forth.
This concept of data-generating environments can be related to FL by considering each client as producing data generated from a different environment. 
However despite this equivalence, existing OOD generalization techniques can't be directly applied to FL. Apart from the FL-specific challenges, these approaches also suffer from several key limitations in non-FL domains \cite{rosenfeld2020risks,nagarajan2020understanding}, further rendering them unfit for practical deployment. Thus developing causal inference models for FL which are inspired from invariant prediction in OOD generalization, are bound to inherit the failures of the latter. 
\par In this work, we consider one such popular OOD generalization technique, \irmgames as it formulation bears close resemblance to a standard FL setup. However, as discussed, \irmgames too suffers from various challenges, which impede its deployment in a generic ML framework, specifically in FL. In the following section, we elaborate on each of these limitations and discuss the corresponding modifications required to overcome them. Further, inspired from the game theoretic formulation of \irmgames, we propose \flgames which forfeits it's failures and can recover the causal mechanisms of the targets, while also providing robustness to changes in the distribution.

\subsection{Challenges in Federated Learning}\label{challenges_fl}
\noindent \textbf{Data Privacy.} Consider a FL system with $m$ client devices, $\mathcal{S}=\{1,2,...,m\}$. Let $N_k$ denote the number of data samples at client device $k$, and $\mathcal{D}_k = \{(x_i^k, y_i^k)\}_{i=1}^{N_k}$ as it's labelled dataset.  The constraint of each environment in \irmgames can be used to formulate the local objective of each client. In particular, each client $k \in \mathcal{S}$ now serves as a player, competing to learn $w^k \in \mathcal{H}_w$ by optimizing its local objective, i.e. $w^k \in \argmin_{w_k' \in \mathcal{H}_w} R^k\Bigg( \frac{1}{|\mathcal{S}|}(w_k' + \sum_{\substack{q \in \mathcal{E}_{tr} \\ q \neq k}} w^q)\circ \phi\Bigg), \forall w_k' \in \mathcal{H}_w$
However, the upper-level objective of \irmgames requires optimization over the dataset pooled together from all environments. Centrally hosting the data at the server or sharing it across clients contradicts the objective of FL. 
Hence, we propose using \fedsgd \cite{mcmahan2017communication} to optimize $\phi$. Specifically, each client $k$ computes and broadcasts $g^k = \nabla R^k\Big( \frac{1}{|\mathcal{S}|}(w_k' + \sum_{q \neq k} w^k)\circ \phi\Big)$ to the server, which then aggregates these gradients and applies the update rule $\phi_{t+1} = \phi_{t} - \eta\sum_{k \in \mathcal{S}}\frac{N_k}{N} g^k$ where $g^k$ is computed over $C$\% batches locally and $N = \sum_{k\in S} N_k$. We call this approach \vfl. The variant with $\phi=I$ is called \ffl. \\


\noindent \textbf{Sequential dependency} \label{algo_sequential}
As discussed, \irmgames poses \irm objective as finding the Nash equilibrium of an ensemble game across environments and adopts the classic best response dynamics (\brd) algorithm to compute it. This approach is based on playing clockwise sequences wherein players take turns in a fixed cyclic order, with only one player being allowed to change its action at any given time $t$ (details in the supplement). 
This linear scaling of time complexity with the number of players poses a major challenge solving the game in FL. 
We modify the classic \brd algorithm by allowing simultaneous updates at any given round $t$. However, a client now best responds to the optimal actions played by it's opponent clients in round $t-1$ instead of $t$. We refer to this approximation of BRD in \ffl and \vfl as \textit{parallelized} \ffl and, \textit{parallelized} \vfl respectively.\\

\noindent \textbf{Oscillations.} As demonstrated in \cite{ahuja2020invariant}, when a neural network is trained using the \irmgames objective (equation \ref{eq:irmgames_bilevel}), the training accuracy initially stabilizes at a high value and eventually starts to oscillate.
The explanation for these oscillations attributes to the significant difference among the data of the training environments. 
\indent With model's performance metrics oscillating to and from at each step, defining a reasonable stopping criterion becomes challenging. As shown in various game theoretic literature \cite{herings2017best,barron2010best,fudenberg1998theory,ge2018fictitious}, \brd can often oscillate. Computing the Nash equilibrium for general games is non-trivial and is only possible for a certain class of games (e.g., concave games) \cite{zhou2017mirror}.  Thus, rather than alleviating oscillations completely, we propose solutions to reduce them significantly to better target valid stopping points. We propose a two-way ensemble approach wherein apart from maintaining an ensemble across clients ($w^{av}$), each client $k$ also responds to the ensemble of historical models (memory) of its opponents. 
Formally, we maintain queues (a.k.a. buffer) at each client which store its historically played actions. In each iteration, a client
best responds to a uniform distribution over the past strategies of its opponents. The global objective at the server remains unchanged. Mathematically, the new local objective of each client $k \in \mathcal{E}_{tr}$ can be stated as

\begin{small}
\begin{equation} \label{eq:fictplay_bilevel}
w^k \in \argmin_{w_k' \in \mathcal{H}_w} R^k\Bigg( \frac{1}{|\mathcal{S}|}(w_k' + \sum_{\substack{q \in \mathcal{S} \\ q \neq k}} w^q + \sum_{\substack{p \in \mathcal{S} \\ p \neq k}} \frac{1}{|\mathcal{B}_p|}\sum_{
j=1}^{ |\mathcal{B}_p|}w^p_j)\circ \phi\Bigg)
\end{equation}
\end{small}

where $\mathcal{B}_q$ denotes the buffer at client $q$ and $w^q_j$ denotes the $j$th historical model of client $q$. 
As the buffer reaches its capacity, it is renewed based on first in first out (FIFO) manner. Note that this approach doesn't result in any communication overhead since a running sum over historical strategies can be calculated in $\mathcal{O}(1)$ time by maintaining a prefix sum.
This variant is called \ffictplay or \vfictplay based on the constraint on $\phi$.\\

\noindent \textbf{Convergence speed.}
As discussed, \flgames has two variants: \ffl and, \vfl with the former being an approximation of the latter ($\phi = I$). 
While both the approaches exhibit superior performance on a variety of benchmarks, the latter has shown its success in a variety of large scale tasks like language modeling. \cite{peyrard2021invariant}. Despite being theoretically grounded, \virm suffers from slower convergence due to an additional round for optimization of $\phi$. 
\par Typically, clients (e.g. mobile devices) posses fast processors and computational resources and have datasets that are much smaller compared to the total dataset size. Hence, utilizing additional local computation is essentially free compared to communicating with the server. To improve the efficiency of our algorithm, we propose replacing the stochastic gradient descent (SGD) over $\phi$ by a full-batch gradient descent (GD).
Intuitively, now at each gradient step, the resultant $\phi$ takes large steps in the direction of its global optimum, resulting in fast training. Note that the classifiers at each client are still updated over one mini-batch. This variant of \flgames is referred to \textsc{V-FL Games (Fast)}.

\section{Experiments and Results} \label{results_section}
In \cite{ahuja2020invariant}, \irmgames was tested over a variety of benchmarks including \coloredmnist, \coloredfashionmnist and \dsprites dataset. We utilize the same datasets for our experiments. Additionally, we create another benchmark, \cifar. Details on each of the datasets can be found in the supplement. We report the mean performance of various baselines over 5 runs. 


\par We compare these algorithmic variants across fixed and variable $\phi$ separately as shown in the Table \ref{tab:combinedresultstable}. Clearly, across all benchmarks, the FL baselines \fedsgd, \fedavg \cite{mcmahan2017communication}, \fedbn \cite{li2021fedbn} and \fedprox \cite{li2020federated} are unable to generalize to the test set, with \fedbn exhibiting superior performance compared to the others. Intuitively, these approaches latch onto the spurious features to make prediction, hence leading to poor generalization over novel clients. From Table \ref{tab:combinedresultstable}, we observe that all modifications in \flgames individually achieve high testing accuracy, hence eliminating the spurious correlations unlike state-of-the-art FL techniques. Further, on \dsprites, we discover that the mean performance of \vfl is superior to all algorithms with fixed representation ($\phi=I$). This accentuates the importance of \vfl over \ffl, especially over complex and larger datasets where learning $\phi$ becomes imperative.



\begin{table*}[!htb]
\centering
\caption{\small{Comparison of methods in terms of training and testing accuracy (mean $\pm$ std deviation).'Seq.' and 'Par.' are abbreviations for sequential and parallel respectively.}}\label{tab:combinedresultstable}
\begin{adjustbox}{width=\textwidth}
\begin{tabular}{l|l|l|p{1.8cm}p{2cm}|p{1.8cm}p{2cm}|p{1.8cm}p{2cm}|p{1.8cm}p{2cm}}
\toprule
& & & \multicolumn{2}{|c|}{\coloredmnist} &  \multicolumn{2}{c|}{\coloredfashionmnist} & \multicolumn{2}{c|}{\cifar} & \multicolumn{2}{c}{\dsprites}\\
\midrule
& &\sharutalgo & \textsc{Train \newline  Accuracy}\xspace  & \textsc{Test \newline  Accuracy}\xspace & \textsc{Train \newline  Accuracy}\xspace & \textsc{Test \newline  Accuracy}\xspace & \textsc{Train \newline  Accuracy}\xspace & \textsc{Test \newline  Accuracy}\xspace & \textsc{Train \newline  Accuracy}\xspace & \textsc{Test \newline  Accuracy}\xspace\\
\midrule
\multirow{2}{*}{\rotatebox[origin=c]{90}{\textbf{Baselines}}} & &\fedsgd & 84.88 $\pm$ 0.16 &  10.45 $\pm$ 0.60 & 83.49 $\pm$ 1.22 & 20.13 $\pm$ 8.06 & 84.79 $\pm$ 0.17 & 12.57 $\pm$ 0.55 & 99.15 $\pm$ 1.10 & 24.12 $\pm$ 2.00\\
& &\fedavg & 84.45 $\pm$ 2.69 &  12.52 $\pm$	4.34& 86.23 $\pm$ 0.63 &	13.33  $\pm$ 2.07 & 85.41 $\pm$ 1.45 & 13.11 $\pm$ 1.82 & 99.21 $\pm$ 1.35 & 22.56 $\pm$ 2.34\\
& & FedBN & 99.75 $\pm$ 0.11 & 47.16 $\pm$ 3.76 & 99.79 $\pm$ 0.17 & 41.24 $\pm$ 1.87 & 95.24 $\pm$ 2.34& 25.16 $\pm$ 4.06 & 98.09 $\pm$ 1.45 & 25.19 $\pm$ 1.78 \\
& & FedPROX & 99.56 $\pm$ 0.38 & 29.31 $\pm$ 0.89 & 99.87 $\pm$ 0.12 & 29.41 $\pm$ 0.36& 99.67 $\pm$ 0.13& 24.76 $\pm$ 2.67  & 85.17 $\pm$ 1.95 & 11.12 $\pm$ 0.99 \\

\midrule[1pt]

\multirow{4}{*}{\rotatebox[origin=c]{90}{\textbf{Fixed}}} & \multirow{2}{*}{\rotatebox[origin=c]{90}{Seq.}} & \ffl & 55.76 $\pm$ 2.03 & 66.56 $\pm$ 1.58 & 75.13 $\pm$ 1.38 &  68.40 $\pm$ 1.83 & 50.36 $\pm$ 2.78 & 45.36 $\pm$ 4.33 & 53.98 $\pm$ 3.67 & 52.89 $\pm$ 4.41\\
& &\ffictplay & 62.83 $\pm$ 5.06 & 66.83 $\pm$ 1.83 & 75.18 $\pm$ 0.37 & \textbf{71.81 $\pm$ 1.60} & 64.02 $\pm$ 2.08& 45.54 $\pm$ 1.04 & 52.87 $\pm$ 3.30 & 61.45 $\pm$ 7.11\\
\cmidrule{2-11}
& \multirow{2}{*}{\rotatebox[origin=c]{90}{Par.}} & \ffl & 58.03 $\pm$ 6.22 & 67.14 $\pm$ 2.95 & 71.71 $\pm$ 8.23 &  69.73 $\pm$ 2.12  & 55.06 $\pm$ 2.04 & 52.07 $\pm$ 1.60 & 52.88 $\pm$ 2.78 & 56.50 $\pm$ 6.23\\
&  & \ffictplay & 61.07 $\pm$ 1.71 & \textbf{67.21 $\pm$  2.98} &72.81 $\pm$ 4.51 &  71.36 $\pm$ 4.19 & 56.98 $\pm$ 4.09 & \textbf{54.71 $\pm$ 2.13} & 53.65 $\pm$ 2.11 & \textbf{62.76 $\pm$ 5.97}\\

\midrule[1pt]

\multirow{4}{*}{\rotatebox[origin=c]{90}{\textbf{Variable}}} & \multirow{2}{*}{\rotatebox[origin=c]{90}{Seq.}} & \vfl & 56.40 $\pm$  0.03 & 63.78 $\pm$ 1.58 & 69.90 $\pm$ 4.56 & \textbf{69.90 $\pm$ 1.31} & 61.72 $\pm$ 7.39 & 46.07 $\pm$ 6.01 & 51.36 $\pm$ 5.32 & 62.84 $\pm$ 7.20 \\
& &\finalalgo & 61.03 $\pm$ 3.11 &  65.81 $\pm$ 3.28 & 75.10  $\pm$ 0.48 & 69.85 $\pm$ 1.22 & 50.37 $\pm$ 4.97 & \textbf{50.94 $\pm$ 3.28} & 51.55 $\pm$ 3.20 & 68.23 $\pm$ 4.56\\
\cmidrule{2-11}
& \multirow{2}{*}{\rotatebox[origin=c]{90}{Par.}} & \vfl & 52.89 $\pm$ 8.03 & \textbf{68.34 $\pm$ 5.24} & 66.33 $\pm$ 9.39 & 69.85 $\pm$ 3.42 & 50.41 $\pm$ 3.31&  50.43 $\pm$ 3.04  & 53.56 $\pm$ 4.91 & 65.87 $\pm$ 6.84\\
& &\finalalgo & 63.11 $\pm$  3.02 & 65.73 $\pm$ 1.53 &  71.89 $\pm$ 5.58 & 69.41 $\pm$ 5.49 & 45.83 $\pm$ 2.44 & 49.89 $\pm$5.66 & 54.25 $\pm$ 2.05  &  \textbf{68.91 $\pm$ 6.47}\\

\midrule[1pt]

& & \textsc{Optimal}\xspace & 75 & 75 & 75 & 75  & 75 & 75 & 75 & 75\\
\bottomrule
\end{tabular}
\end{adjustbox}
\end{table*}

\par In all the above experiments, both of our end approaches: \textit{parallelized \finalalgo} and \textit{parallelized} \ffictplay are able to perform better than or at par with the other variants. \textit{These algorithms were primarily designed to overcome the challenges faced by causal FL systems while retaining their original ability to learn causal features.} Hence, the benefits provided by these approaches in terms of 1) robust predictions; 2) scalability; 3) fewer oscillations and 4) fast convergence are not at the cost of performance. While the former is demonstrated by Table \ref{tab:combinedresultstable}, the latter three are detailed in Section \ref{ablation}.

\subsection{Interpretation of Learned Features} \label{lime_interpret}
\begin{figure}[htb!]
\centering
  \includegraphics[width=1.0\textwidth]{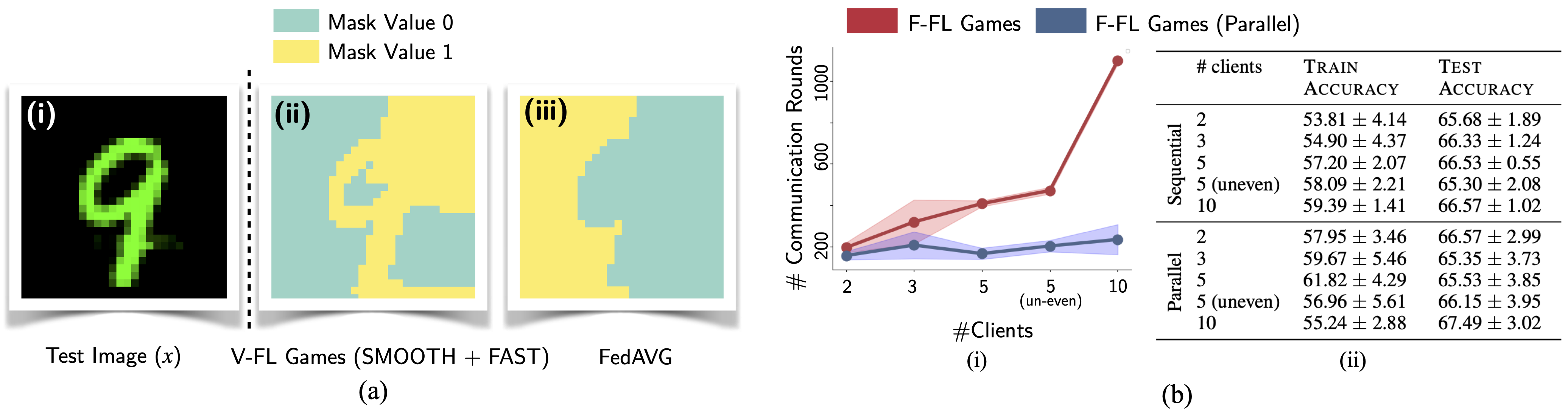}
\caption{\small{(a)(i) Test image used as input to the interpretable model; LIME mask corresponding to model prediction using (ii)\finalalgo; (iii)\fedavg; (b) Comparison of \ffl and \ffl (Parallel) with increasing client in terms of (i) communication rounds; (ii) performance.}}
\label{fig:fig_maintext1}
\end{figure}

In order to explain the predictions of our parametric model, we use LIME \cite{ribeiro2016should} to learn an interpretable model around each prediction. Figure \ref{fig:fig_maintext1}(a) shows LIME masks for both \fedavg (ii) and our method (iii). Clearly, the former focuses solely on the background to make the model prediction. However, the latter uses causal features in the image (like shape, stroke, edges and curves) along with some pixels from the background (noise) to make predictions. This demonstrates the reasoning behind robust generalization of FL GAMES as opposed to state-of-the-art FL techniques.

\subsection{Ablation Analysis} \label{ablation}
We analyze the effect of each of our algorithmic modifications on the \coloredmnist dataset. The results on other datasets are similar and are provided in the supplement.

\subsubsection{Effect of Simultaneous \brd} \label{parallel_ablation}
We examine the effect of replacing the classic best response dynamics as in \cite{ahuja2020invariant} with the simultaneous best response dynamics. For the same, we use a more practical environment: (a) more clients are involved, and (b) each client has fewer data. Similar to \cite{choe2020empirical}, we extended the \coloredmnist dataset by varying the number of clients between 2 and 10. For each setup, we vary the degree of spurious correlation between 70\% and 90\%) for training clients and merely 10\% in the testing set.  A more detailed discussion of the dataset is provided in the supplement. For \ffl, it can be observed from Figure \ref{fig:fig_maintext1}(b.i), as the number of clients in the FL system increase, there is a sharp increase in the number of communication rounds required to reach equilibrium. However, the same doesn't hold true for \textit{parallelized} \ffl. Further,  \textit{parallelized} \ffl is able to reach a comparable or higher test accuracy as compared to \ffl with significantly lower communication rounds (refer to Figure \ref{fig:fig_maintext1}(b.ii)).

\subsubsection{Effect of a memory ensemble} \label{fictplay_ablation}
\begin{figure}[!htb]
\centering
  \includegraphics[width=1.0\textwidth]{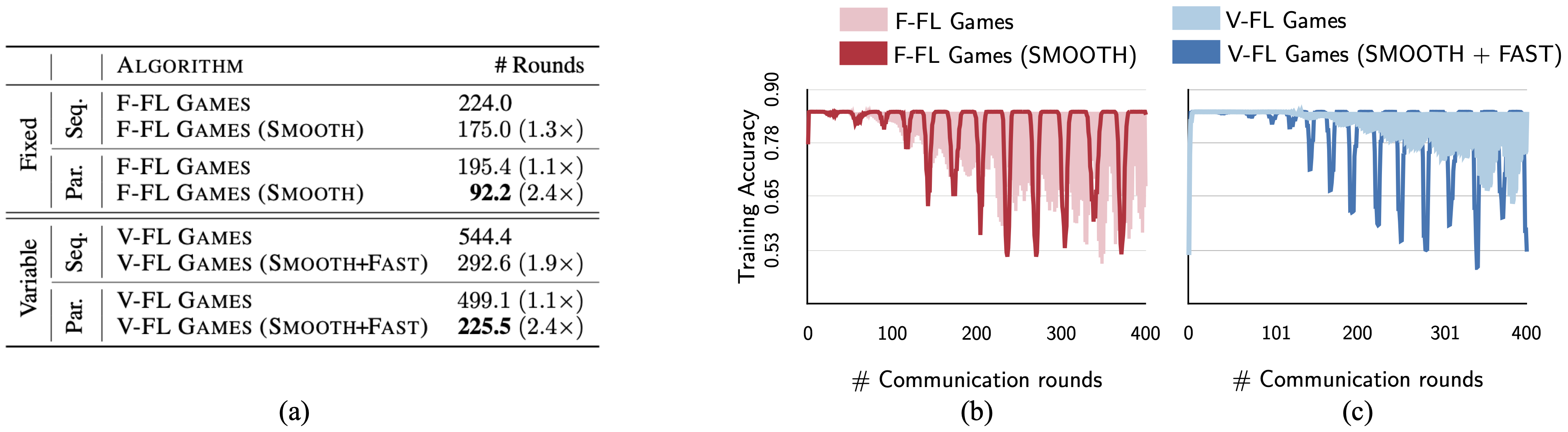}
\caption{\small{\coloredmnist: (a) Comparison of \ffl and \ffl (Parallel) with increasing clients; Training accuracy of (b) \ffl and \ffictplay for a buffer size of 5; (c) \vfl and \finalalgo with buffer size as 5 versus the number of rounds;}}
\label{fig:fig_maintext2}
\end{figure}

As shown in Figure \ref{fig:fig_maintext2}(b), compared to \ffl, \ffictplay reduces the oscillations significantly. While in the former, performance metrics oscillate at each step, the oscillations in the later are observed after an interval of roughly 50 rounds. Further, \ffictplay seems to envelop the performance curves of \ffl. As a result, apart from reducing the frequency of oscillations, \ffictplay also achieves higher testing accuracy compared to \ffl. The observations are consistent across the \textit{parallelized} variants. Further, as observed from Figure \ref{fig:fig_maintext2}(a), both sequential and parallel \ffictplay significantly reduce the number of rounds required to reach equilibrium, further underscoring the efficacy of our proposed methodology.

\subsubsection{Effect of using Gradient Descent (GD) for $\phi$}
Communication costs are the principal constraints in FL setup. Edge devices like mobile phones and sensor are bandwidth constrained and require more power for transmission and reception as compared to remote computation. As observed from Figure  \ref{fig:fig_maintext2}(c), \finalalgo is able to achieve significantly higher testing accuracy in fewer rounds as compared to \vfl. Consistent results are also reported in Figure \ref{fig:fig_maintext2}(a), where both sequential and parallel variants of \finalalgo result in a significant improvement ($\sim$2×) in the number of rounds.

\section{Conclusion}
In this work, we develop a novel framework based on Best Response Dynamics (\brd) training paradigm to learn invariant predictors across clients in Federated learning (FL). Inspired from \cite{ahuja2020invariant}, the proposed method called Federated Learning Games (\flgames) learns causal representations which have good out-of-distribution generalization on new train clients or test clients unseen during training. We investigate the high frequency oscillations observed using \brd and equip our algorithm with a memory of historical actions. This results in smoother performance metrics with significantly lower oscillations. \flgames exhibits high communication efficiency as it allows parallel computation, scales well in the number of clients and results in faster convergence. Given the impact of FL in medical imaging, we plan to test our framework over medical benchmarks. Future directions include theoretically analyzing the smoothed best response dynamics, as it might have potential implications for other game-theoretic based machine learning frameworks. 


\clearpage
\bibliographystyle{plainnat}
\bibliography{aaai22}

\clearpage
\appendix

\section{Appendix}

\section{Game Theory Concepts} \label{background_gametheory}

We define some basic game theory notations that will be used later. Let $\Gamma = (N, \{S_k\}_{k \in N}, \{u_k\}_{k \in N})$ be the tuple representing a normal form game, where $N$ denotes the finite set of players.  For each player $k$, $S_k = \{s^k_0, s^k_1,...s^k_m\}$ denotes the pure strategy space with $m$ strategies and $u_k(s_k, s_{-k})$ denotes the  payoff function of player $k$ corresponding to strategy $s_k$. Here, an environment for player $k$ is $s_{-k}$, a set containing strategies taken by all players but $k$ and $S_{-k}$ denotes the space of strategies of the opponent players to $k$. $S = \prod_{i \in N} S_i$ denotes the joint strategy set of all players. A game $\Gamma$ is said to be finite if $S$ is finite and is continuous if $S$ is uncountably infinite.
\par While a pure strategy defines a specific action to be followed at any time instance, a mixed strategy of player $k$, $\sigma_k = \{p_k(s^k_0), p_k(s^k_1), ... p_k(s^k_m)\}$ is a probability distribution over a set of pure strategies, where $\sum_{j=1}^m p_k(s^k_j)=1$. The expected utility of a mixed strategy $u_k(\sigma_k, \sigma_{-k})$ for player $k$ is the expected value of the corresponding pure strategy payoff i.e.
\begin{equation}
\begin{aligned}
\mathbb{E}(u_k(\sigma_k, \sigma_{-k})) = \sum_{s_k \in S_k} \sum_{s_{-k} \in S_{-k}} u_k(s_k, s_{-k})p_k(s_k) p_{-k}(s_{-k}), \forall \sigma_k \in \Tilde{S}_k
\end{aligned}
\end{equation}
where $\Tilde{S}_k$ corresponds to the mixed strategy space of player $k$.

\noindent \textbf{Best response (BR).} A mixed strategy $\sigma^*_k$ for player $k$ is said to be a best response to it's opponent strategies $\sigma_{-k}$ if 
$$ \mathbb{E}(u_k(\sigma^*_k, \sigma_{-k})) \geq \mathbb{E}(u_k(\sigma_k, \sigma_{-k})), \forall \sigma_k \in \Tilde{S}_k.$$

\noindent \textbf{Nash equilibrium.} A mixed strategy profile $\sigma^* = \{\sigma^*_1,\sigma^*_2,...\sigma^*_N\}$ is a Nash equilibrium if for all players $k$, $\sigma^*_k$ is the best response to the strategies played by it's opponent players i.e $\sigma^*_{-k}$.\\

\noindent \textbf{Best response dynamics (\brd).}  \brd is an iterative algorithm in which at each time step, a player myopically plays strategies that are best responses to the most recent known strategies played by it's opponents previously. Based on the playing sequence across layers, \brd can be classified into three broad categories: \brd with clockwise sequences, \brd with simultaneous updating and \brd with random sequences. For this study, we focus only on the first two playing schedules. Let the function $\text{seq}: \mathbb{N} \rightarrow \mathcal{P}[N]$ denote a playing sequence which determines the set of players whose turn it is to play at each time period $t \in \mathbb{N}$. Here, $\mathcal{P}[N]$ denotes the power set of $\{1,2,...N\}$ players and $\mathbb{N}$ be the set of natural numbers $\{1,2,...\}$. By \brd, at each time step $t$, $\forall i \in \text{seq}(t)$, action taken by player $i$ i.e. $a_i^t$ is the best response to it's current environment i.e $a_{-i}^t$. 
\begin{itemize}
    \item \brd with clockwise sequences: In this playing sequence, players take turns according to a fixed cyclic order and only one player is allowed to change it's action at any given time $t$. Specifically, the playing sequence is defined by $\text{seq}(t) = 1 + (t-1) \mod n$. Since only a single player is allowed to play at any given time $t$, $a_{\text{-seq}(t)}^t = a_{\text{-seq}(t)}^{t-1}$. 
    \item \brd with simultaneous updating: In this playing sequence, $\text{seq}(t)$ chooses a non empty subset of players to participate in round $t$. However, for each player $i \in \text{seq}(t)$, the optimal action chosen $a_i^t$ depends on the knowledge of the latest strategy of it's opponents.
\end{itemize}

\section{Datasets}
The MNIST dataset consists of handwritten digits, with a total of 60,000 images in the training set and 10,000 images in the test set \cite{deng2012mnist}. These images are black and white in colour and form a subset of the larger collection of digits called NIST. Each digit in the dataset is normalized in size to centre fit in the fixed size image of size 28$\times$28. It is then anti-aliased to introduce appropriate gray-scale levels. 

\subsection{\coloredmnist}
\begin{figure}[htb!]
    \centering
\includegraphics[width=.8\columnwidth]{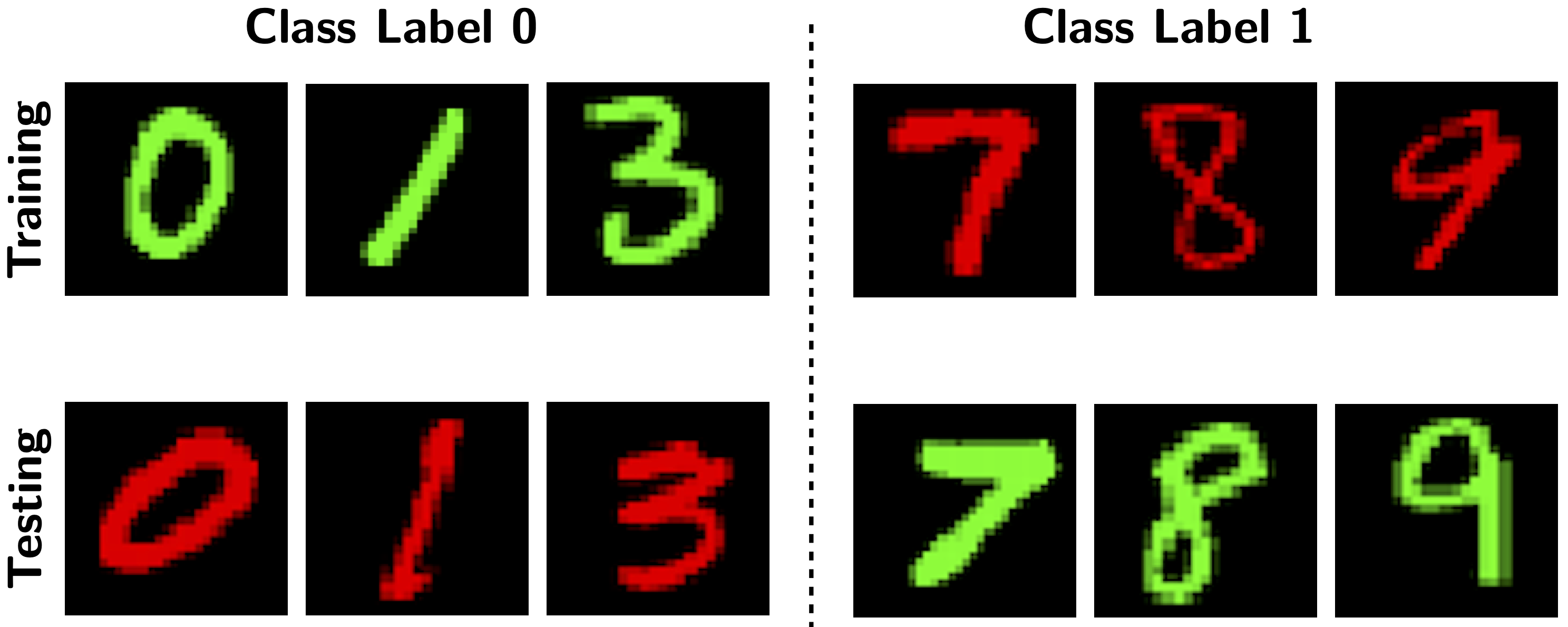}
    \caption{\small{\coloredmnist: Illustration of samples containing high spurious correlation between label and color during training. While testing, this correlation is significantly reduced as label 0 is highly correlated with the color mapped to label 1 and vice-versa.}}
    \label{fig:colored_mnist_dataset}
\end{figure}

We modify the MNIST dataset in the exact same manner as in \cite{arjovsky2019invariant}. Specifically, \cite{arjovsky2019invariant} creates the dataset in a way that it contains both the invariant and spurious features according to different causal graphs. Spurious features are introduced using colors. Digits less than 5 (excluding 5) are attributed with label 0 and the rest with label 1. The dataset is divided across three clients, out of which two serve as training and one as testing. The 60,000 images from MNIST train set are divided equally amongst the two training clients i.e. each consists of 30,000 samples. The testing set contains the 10,000 images from the MNIST test set. Preliminary noise is added to the label to reduce the invariant correlation. Specifically, the initial label ($\Tilde{y}$) of each image is flipped with a probability $\delta_k$ to construct the final label $y$. The final label $y$ of each image is further flipped with a probability $p_k$ to construct it's color code ($z$). In particular, the image is colored red, if $z=1$ and green if $z=0$.
The flipping probability which defines the color coding of an image, $p_k$ is 0.2 for client 1, 0.1 for client 2 and 0.9 for the test client. The probability $\delta_k$ is fixed to 0.25 for all clients $k$. The above choice is defined in a way that the mean degree of label-color (spurious) correlation ($1-p_k, \forall k$) is more than the average degree of invariant correlation ($1-\delta_k, \forall k$). A sample batch of images elucidating the above construction is shown in Figure \ref{fig:colored_mnist_dataset}.

\subsection{\coloredfashionmnist}
\begin{figure}[htb!]
    \centering
\includegraphics[width=.8\columnwidth]{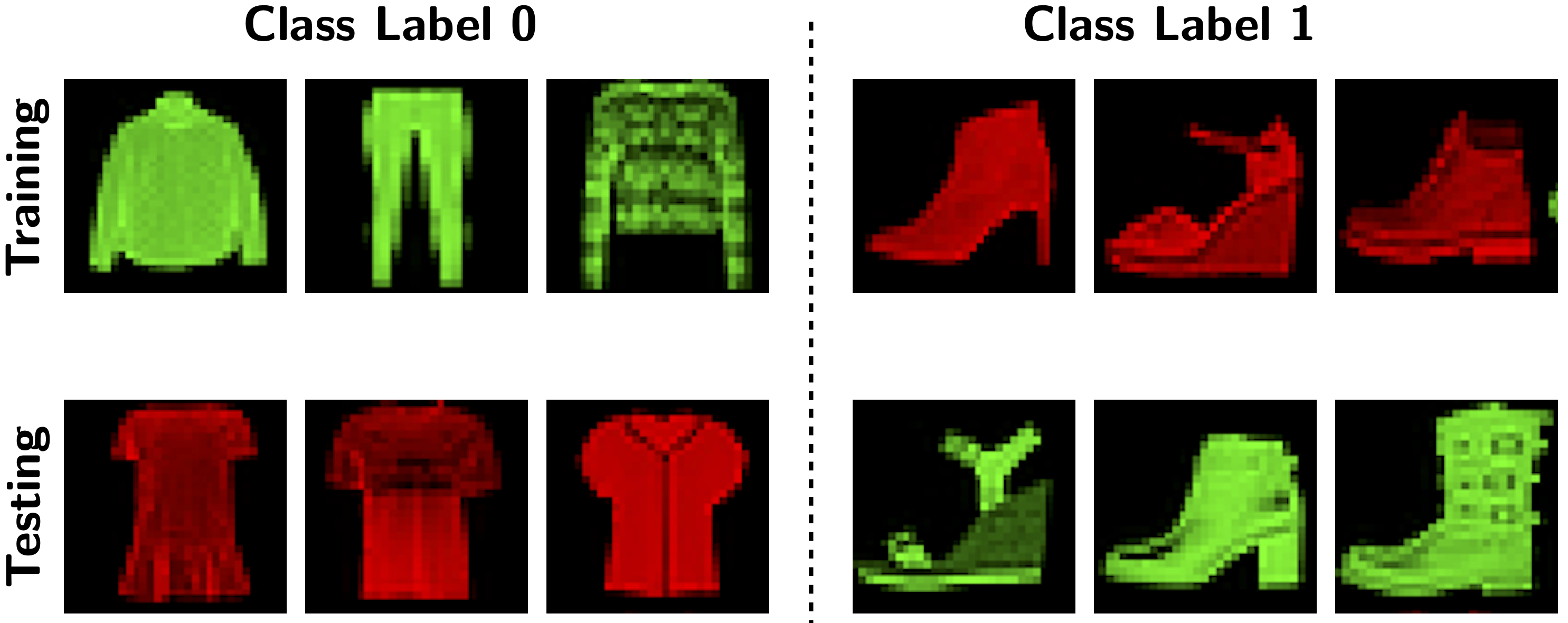}
    \caption{\small{\coloredmnist: Illustration of samples containing high spurious correlation between label and color during training. While testing, this correlation is significantly reduced as label 0 is highly correlated with the color mapped to label 1 and vice-versa.}}
    \label{fig:colored_fashion_mnist_dataset}
\end{figure}

We use the exact same environment for creating \coloredfashionmnist as in \cite{ahuja2020invariant}. The data generating process of \coloredfashionmnist is motivated from that of \coloredmnist in a way that it possesses spurious correlations between the label and the colour. Fashion MNIST consists of images from a variety of sub-categories under the two broad umbrellas of clothing and footwear. Clothing items include categories like: ``t-shirt", ``trouser", ``pullover", ``dress", ``shirt" and ``coat" while the footwear category includes ``sandal", ``sneaker", ``bag" and ``ankle boots". Similar to \coloredmnist, the train dataset is equally split across two clients (30,000 images each) and the entire test set is attributed to the test client. Preliminary labels for binary classification are constructed such  $\Tilde{y} = 0$ for “t-shirt”, “trouser”, “pullover”, “dress”, “coat”, “shirt” and $\Tilde{y} = 1$: “sandle”, “sneaker” and  “ankle boots”. Next, we add noise to the preliminary label by flipping $\Tilde{y}$ with a probability $\delta_k = 0.25, \forall k$ to construct the final label $y$. We next flip the final label with a probability $p_k$ to designate a color ($z$), with $p_1 = 0.2$ for the first client, $p_2 = 0.1$ for the second client and $p_3 = 0.9$ for the test client. The image is colored red, if $z=1$ and green if $z=0$. A sample batch of images elucidating the above construction is shown in Figure \ref{fig:colored_fashion_mnist_dataset}.

\subsection{\cifar} 
\begin{figure}[htb!]
    \centering
\includegraphics[width=.8\columnwidth]{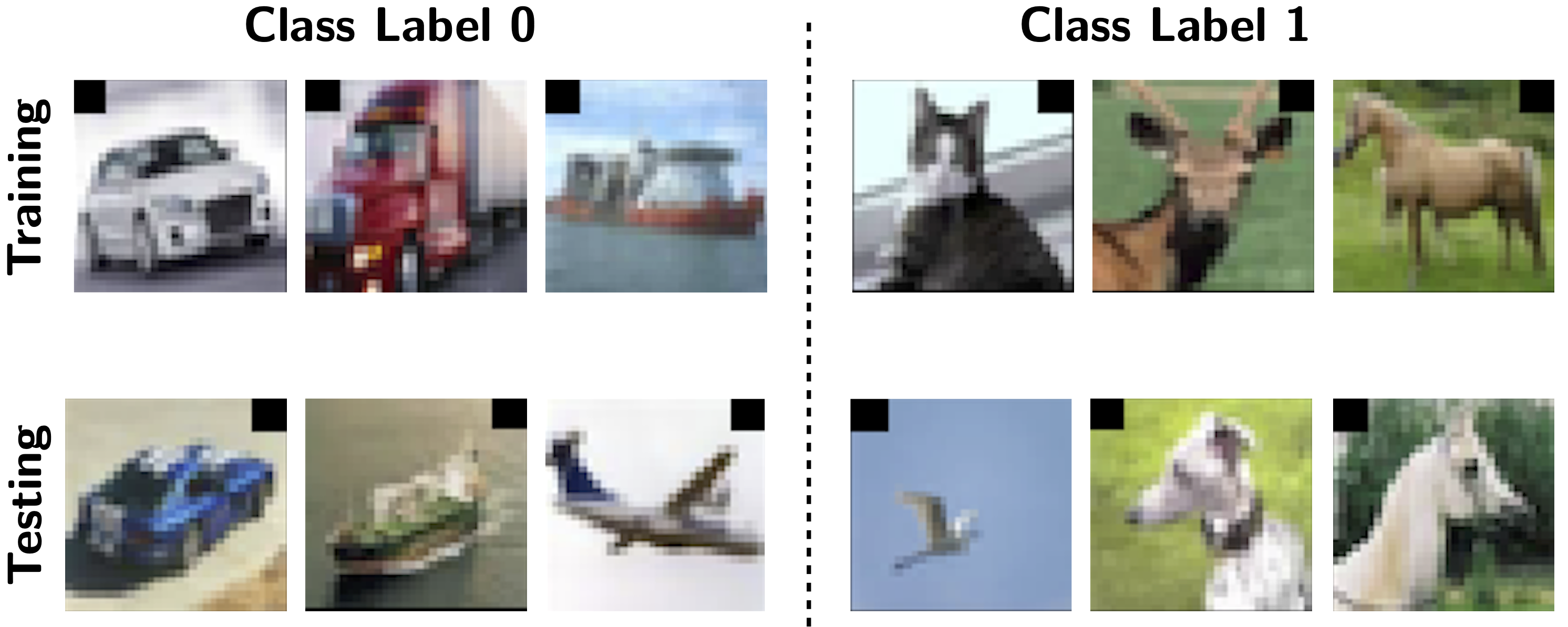}
    \caption{\small{\cifar: Illustration of samples containing high spurious correlation between labels and spatial location of the 5$\times$5 black patch during training. While testing, this correlation is significantly reduced as label 0 is now highly correlated with the spatial location corresponding to label 1 and vice-versa.}}
    \label{fig:spurious_cifar10_dataset}
\end{figure}

In this setup, we modify the CIFAR-10 dataset similar to the \coloredmnist dataset. Instead of coloring the images, we use a different mechanism based on the spatial location of a synthetic feature to generate spurious features. CIFAR10 dataset consists of 60,000 images from 10 classes including ``airplane", ``automobile", ``bird", ``cat", ``deer", ``dog", ``frog", ``house", ``ship", ``truck". The original dataset is relabelled to create a binary classification task between motor and non-motor objects. All images corresponding to the label ``frog" are discarded to ensure a similar samples count for the two classes. Similar to \coloredmnist, the train dataset is equally split across two clients and the entire test set is attributed to the test client. Preliminary labels for binary classification are constructed such  $\Tilde{y} = 0$ for “airplane”, “automobile”, “ship”, “truck” and $\Tilde{y} = 1$: “bird”, “cat”, ``deer",``dog" and  “horse”. Next, we add noise to the preliminary label by flipping $\Tilde{y}$ with a probability $\delta_k = 0.25, \forall k$ to construct the final label $y$. We next flip the final label with a probability $p_k$ to designate a positional index ($z$), with $p_1 = 0.2$ for the first client, $p_2 = 0.1$ for the second client and $p_3 = 0.9$ for the test client. This index defines the spatial location of a 5$\times$5 black patch in the image. An index value of 0 ($z=0$) specifies the patch at the top left corner of the image, while an index value of 1 ($z=1$) corresponds to a black patch over the top-right corner of the image. Based on the choice of flipping probabilities, images in the training set are spuriously correlated to the position of the patch in the image. Such correlation does not exist while testing. A sample batch of images elucidating the above construction is shown in Figure \ref{fig:spurious_cifar10_dataset}.

\subsection{\dsprites}
\begin{figure}[htb!]
    \centering
\includegraphics[width=.8\columnwidth]{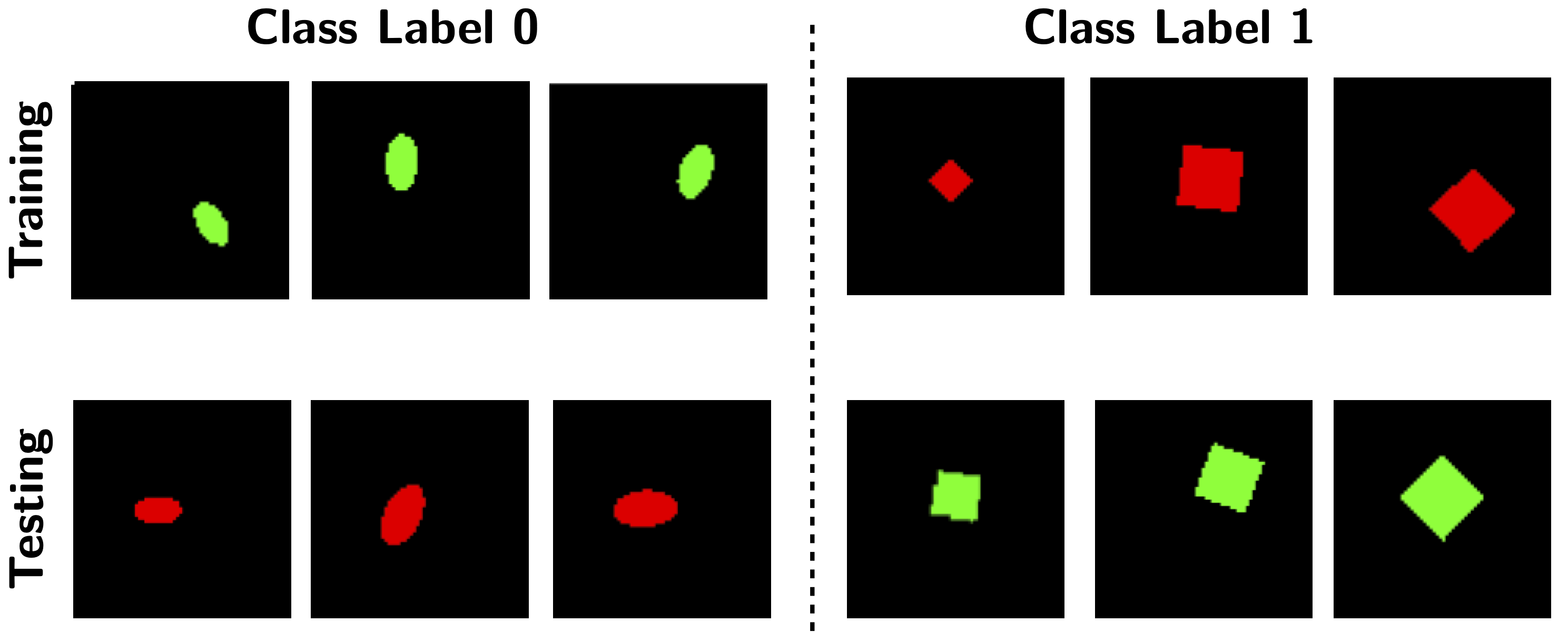}
    \caption{\small{\cifar: Illustration of samples containing high spurious correlation between labels and color during training. While testing, this correlation is significantly reduced as label 0 is now highly correlated with the color corresponding to label 1 (while training) and vice-versa.}}
    \label{fig:desprites_dataset}
\end{figure}

To construct this FL setup, we modify the Desprites dataset\footnote{https://github.com/deepmind/dsprites-dataset}(737,280 images) as in\cite{ahuja2020invariant}. The data generating process of \dsprites is motivated from that of \coloredmnist in a way that it possesses spurious correlations between the label and the colour. Desprites dataset consists of 2D shapes procedurally generated from 6 ground truth independent latent factors. These factors are color, shape, scale, rotation, x and y positions of a sprite.
The shape further belongs to a set of three different categories: square, ellipse or heart. We convert this multi-label and multi-class task into binary classification problem. Specifically, we consider classification between a circle and a square. 
 Similar to \coloredmnist, the train dataset is equally split across two clients and the entire test set is attributed to the test client. Preliminary labels for binary classification are constructed such  $\Tilde{y} = 0$ for circle and $\Tilde{y} = 1$ for square. Next, we add noise to the preliminary label by flipping $\Tilde{y}$ with a probability $\delta_k = 0.25, \forall k$ to construct the final label $y$. We next flip the final label with a probability $p_k$ to designate a color ($z$), with $p_1 = 0.2$ for the first client, $p_2 = 0.1$ for the second client and $p_3 = 0.9$ for the test client. The image is colored red, if $z=1$ and green if $z=0$. A sample batch of images elucidating the above construction is shown in Figure \ref{fig:desprites_dataset}.

\subsection{Extended Datasets} \label{multiclassdatasets}
We extend the two datasets \coloredmnist, and \coloredfashionmnist as in \cite{choe2020empirical} to robustly test our approach across multiple clients. Analogous to \coloredmnist with two training environments ($N=2$), we extend the datasets to incorporate $N=2,3,5$ and $10$ training clients. In particular, we attribute each client with a unique flipping probability $p_k$. For each value of $N$, the maximum value of $p_k, \forall k$ is 0.3, while the minimum value is 0.1. The values of $p_k$ for each client are spaced evenly between this range. For example, for the case of $N=5$ clients, the flipping probabilities of clients are $p_1 = 0.3$, $p_2 = 0.2$, and $p_3 = 0.1$. The flipping probability $\delta_k$ which decides the final label $y$ is fixed to 0.25 for all clients. The maximum and the minimum values of $p_k$ are chosen in a way that the average spurious correlation which is 0.8 is more than the invariant correlation i.e. 0.75.
\par
\begin{figure}[htb!]
    \centering
\includegraphics[width=1.0\linewidth]{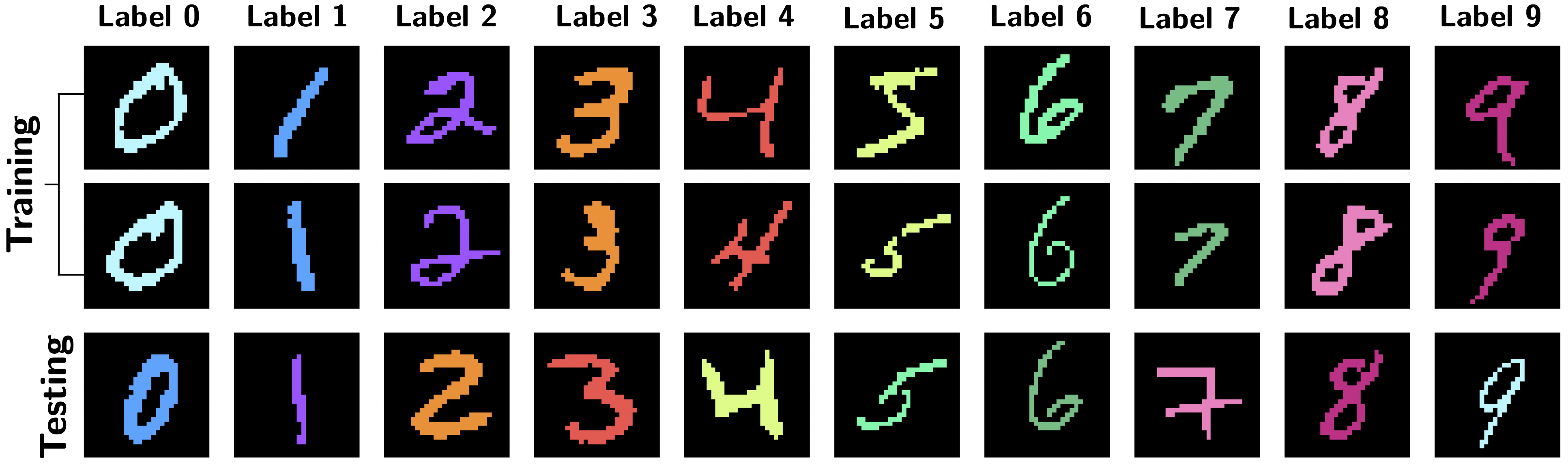}
    \caption{\small{\extendedcoloredmnist: Illustration of samples corresponding to a 10-digit classification task. During training, each class label is spuriously correlated with a unique color. While testing, this correlation is significantly reduced as each image is colored with the color corresponding to its succeeding label. }}
    \label{fig:multiclass_colored_mnist_dataset}
\end{figure}

 Further, since all the previous settings were binary classification tasks,  we extend the standard datasets \coloredmnist and \coloredfashionmnist over multi class classification \cite{choe2020empirical}. Specifically, we extend the number of classes from 2 to 5 and 10. Specifically, a unique color is assigned to each output class such that the label is highly correlated ($\sim 80\% - 90\%$) with the color in the training set. In the test set, these correlations are significantly reduced ($10\%)$ by allowing high spurious correlations with the color of the following class. For instance, in testing, the color corresponding to class label 8 would be the one which was heavily corrected with class label 9 while training. This reduces the original spurious correlations and is hence useful for evaluating the extent to which the trained model has learned the invariant features. A sample batch of images elucidating the above construction is shown in Figure \ref{fig:multiclass_colored_mnist_dataset}.
\par We do not construct a multi-class classification setup for \cifar and \dsprites. For the former, it is difficult to find a unique spatial location in the image corresponding to each class (e.g. in case of 10-class classification). For the latter, the Desprites dataset can be classified into three categories on the basis of shape present in the image:  square, ellipse, and heart. Hence, introducing more than three classes requires categorization of other latent factors like orientation, scale or position.

\section{Experimental Setup}
\subsection{Architecture Details}
For all the approaches using a fixed representation i.e. \ffl, \ffictplay, \textit{parallelized} \ffl and \textit{parallelized} \ffictplay, we use the architecture mentioned below. 
The architecture used to train a predictor at each client is a multi-layered perceptron with three fully connected (FC) layers. The details of layers are as follows:
\begin{itemize}
    \item Flatten: A flatten layer that converts the input of shape \texttt{(Batch Size, Length, Width, Depth)} into a tensor of shape \texttt{(Batch Size, Length * Width * Depth)}
    \item FC1: A fully connected layer with an output dimension of 390, followed by \texttt{ELU} non-linear activation function
    \item FC2: A fully connected layer with an output dimension of 390, followed by \texttt{ELU} non-linear activation function
    \item FC3: A fully connected layer with an output dimension of 2 (Classification Layer)
\end{itemize}
This same architecture is used across approaches with fixed representation.
\par For the approaches with trainable representation i.e. \vfl, \vfictplay, \textit{parallelized} \vfl and \textit{parallelized} \vfictplay,  we use the following
architecture for the representation learner 
\begin{itemize}
    \item Flatten: A flatten layer that converts the input of shape \texttt{(Batch Size, Length, Width, Depth)} into a tensor of shape \texttt{(Batch Size, Length * Width * Depth)}
    \item FC1: A fully connected layer with an output dimension of 390, followed by \texttt{ELU} non-linear activation function
\end{itemize}

The output from FC1 is fed into the following architecture, which is used as the base network to train the predictor at each client:
\begin{itemize}
    \item FC1: A fully connected layer with an output dimension of 390, followed by \texttt{ELU} non-linear activation function
    \item FC2: A fully connected layer with an output dimension of 390, followed by \texttt{ELU} non-linear activation function
    \item FC3: A fully connected layer with an output dimension of 2 (Classification Layer)
\end{itemize}
This architecture is used across approaches with variable representation.

\subsection{Optimizer and other hyperparameters}
We use a different set of hyper-parameters based on the dataset. In particular, \\
\begin{itemize}
    \item \coloredmnist and \coloredfashionmnist: For fixed representation, we use Adams optimizer with a learning rate of 2.5e-4 across all experiments (sequential or parallel). For the variable representation, we use Adams optimizer with a learning rate of 2.5e-5 for the representation learner and the same optimizer with a learning rate of 2.5e-4 for the predictor at each client. 
    \item \cifar: For fixed representation, we use Adams optimizer with a learning rate of 1.0e-4 across all experiments (sequential or parallel). For the variable representation, we use Adams optimizer with a learning rate of 9.0e-4 for the representation learner and the same optimizer with a learning rate of 1.0e-4 for the predictor at each client.
    \item \dsprites: For fixed representation, we use Adams optimizer with a learning rate of 9.5e-4 across all experiments (sequential or parallel). For the variable representation, we use Adams optimizer with a learning rate of 2.5e-4 for the representation learner and the same optimizer with a learning rate of 9.0e-4 for the predictor at each client
    
\end{itemize}

For all the experiments, we fix the batch size to 256 and optimize the Cross Entropy Loss. We use the same termination criterion as in \cite{ahuja2020invariant}.  Specifically, we stop training when the observed oscillations become stable and the ensemble model is in a lower training accuracy state. We choose a training threshold and terminate the training as soon as the training accuracy drops below this value. In order to ensure stability of oscillations, we set a period of warm start. In this period, the training is not stopped even if the accuracy drops below the threshold. For variable representation, the duration of this warm start period is set to the number of training steps in an epoch i.e. (training data size/ batch size). However, for the approaches with fixed representation, this period is fixed to $N$ rounds where $N$ is the number of training clients. In particular, for two clients, the warm start period ends as soon as the second client finishes playing its optimal strategy for the first time. \\

\subsection{Computing Environment}
The experiments were done on an Intel Xeon E5 Processor with 2133MHz DDR4 and NVIDIA Tesla v100, 32GB GPU.

\section{Algorithm}
Algorithm \ref{alg:main_algorithm} represents the pseudo code for \finalalgo. In the following analysis, the terms ‘Sequential’ and ‘Parallel’ denote BRD with clockwise playing sequences and simultaneous updates respectively (Lines 20 and 26 of Algorithm 1). We use \flgames as an umbrella term that constitutes all the discussed algorithmic modifications. \ffl and \vfl refer to the privacy preserving variants of \flgames. The approach used to smoothen out the oscillations (Line 4 of Algorithm 1) is denoted by \ffictplay or \vfictplay depending on the constraint on $\phi$. The fast variant with high convergence speed is typified as \finalalgo) (Line 9 of Algorithm 1).
\begin{algorithm}[htb!]
\small
\caption[\textit{Parallelized} \textsc{FL Games (Smooth \!\!+ \!\!Fast)}\xspace]{\textit{Parallelized} \textsc{FL Games (Smooth \!\!+ \!\!Fast)}\xspace}
\label{alg:main_algorithm}
\begin{algorithmic}[1]
\State \textbf{Notations: } $\mathcal{S}$ is the set of $N$ clients;  $\mathcal{B}_k$ and $\mathcal{P}_k$ denote the buffer and information set containing copies of $\mathcal{B}_i,\forall i\neq k \in \mathcal{S}$ at client, $k$ respectively.
\State \textbf{PredictorUpdate(k):}
    \Indent 
    \State \Comment{Two-way ensemble game to update predictor at each client $k$}
    \State $w_k \gets$ SGD$\Big[\ell_k \Big\{ \frac{1}{|\mathcal{E}_{tr}|}(w_k' + \sum_{\substack{q \in \mathcal{E}_{tr} \\ q \neq k}} w^q +$\colorbox{red!25}{$\sum_{\substack{p \in \mathcal{E}_{tr} \\ p \neq k}} \frac{1}{|\mathcal{B}_p|} \sum_{j=1}^{ |\mathcal{B}_p|}w^p_j$}$)\circ \phi\Big\}\Big]$ \label{lst:line:oscillations}
    \State Insert $w_k$ to $\mathcal{B}_k$, discard oldest model in $\mathcal{B}_k$ if full
    \State return $w_k$
\EndIndent
\State \textbf{RepresentationUpdate(k):}
    \Indent 
    \State \Comment{Gradient Descent (GD) over entire local dataset at client $k$}
    \For{\colorbox{yellow!35}{every batch $b \in \mathcal{B}$}} \label{lst:line:fast}
        \State Compute $\nabla \ell_k(w_{\text{cur}}^{\text{av}} \circ \phi_{\text{cur}};b)$; Add in $\nabla\phi_k$
    \EndFor
    \State return $\nabla\phi_k$
\EndIndent
\State \textbf{Server executes:}
\Indent
    \State Initialize $w_k, \forall k \in \mathcal{S}$ and $\phi$  
    \While{round $\leq$ max-round}   
      \State \Comment{Update representation $\phi$ at even round parity}
      \If{round is even} 
        \If{Fixed-Phi}
            \State $\phi_{\text{cur}} = I$
        \EndIf
      
        \If{Variable-Phi} 
            \For{each client $k \in \mathcal{S}$ \colorbox{green!30} {in parallel}} \label{lst:line:parallel1}
                \State $\nabla \phi_k=$ RepresentationUpdate(k)
            \EndFor
            \State \Comment{Update representation $\phi$}
            \State $\phi_{\text{next}} = \phi_{\text{cur}} - \eta\Big(\sum_{k \in \mathcal{S}}\frac{N_k}{\sum_{j \in \mathcal{S}}N_j}  \nabla \phi_k \Big)$ 
            \State $\phi_{\text{cur}} = \phi_{\text{next}}$
        \EndIf
      
      \Else
        \For{each client $k \in \mathcal{S}$ \colorbox{green!30}{in parallel}} \label{lst:line:parallel2}
            \State $w_k$ $\gets$ PredictorUpdate(k)
        \EndFor
        \State \Comment{Client $k$ updates its information set  $\mathcal{P}_k$ by updating copies of predictors of other clients}
        \State Communicate $\forall k, \mathcal{P}_k \gets \{w_i, \forall i\neq k \in S \}$
        
      \EndIf
     \State round $\gets$ round + 1
     \State $w_{\text{curr}}^{\text{av}} = \frac{1}{N} \sum_{k \in \mathcal{S}} w_{\text{curr}}^k $
    \EndWhile
\EndIndent
\end{algorithmic}
\end{algorithm}

\section{Additional Results and Analysis}
\subsection{Robustness to the number of Outcomes}

\begin{table}[htb!]
\caption{\small{\coloredmnist:  Comparison of \ffl and \ffl (Parallel) with increasing number of output classes, in terms of the training and testing accuracy (mean $\pm$ std deviation). Here `Seq.' is an abbreviation used for `Sequential', which denotes \ffl.}}
\centering
\begin{adjustbox}{width=0.55\columnwidth}
\begin{tabular}{llll}
\toprule
 Type & \# Classes & \trainacc & \testacc  \\
\midrule
\multirow{3}{*}{\rotatebox[origin=c]{90}{Seq.}} &	2	& 75.13	$\pm$	1.38 &68.40	$\pm$ 1.83	\\
&	5	&	79.39	$\pm$	0.91	&	69.90	$\pm$	3.16	\\
&	10	&	82.27	$\pm$	0.76	&	69.22	$\pm$	3.11	\\
\midrule
\multirow{3}{*}{\rotatebox[origin=c]{90}{Parallel}} &	2	& 71.71 $\pm$ 8.23 &	69.73 $\pm$ 2.12 \\
&	5	&	78.61	$\pm$	2.86	&	68.42	$\pm$	2.54	\\
&	10	&	82.17	$\pm$	1.21	&	69.29	$\pm$	3.17	\\
\bottomrule
\end{tabular}
\end{adjustbox}
\label{tab:multiple_outcomes_mnist}
\end{table}

\begin{table}[htb!]
\caption{\small{\coloredfashionmnist: Comparison of \ffl and \ffl (Parallel) with increasing number of output classes, in terms of the training and testing accuracy (mean $\pm$ std deviation). Here `Seq.' is an abbreviation used for `Sequential', which denotes \ffl.}}
\centering
\begin{adjustbox}{width=0.55\columnwidth}
\begin{tabular}{llll}
\toprule
 Type & \# Classes & \trainacc & \testacc  \\
\midrule
\multirow{3}{*}{\rotatebox[origin=c]{90}{Seq.}} &	2	&	50.36 $\pm$	 2.78		&	47.36 	$\pm$	 4.33	\\
&	5	&	77.28	$\pm$	1.54	&	69.35	$\pm$	0.66	\\
&	10	&	80.02	$\pm$	0.38	&	71.22	$\pm$	3.00	\\
\midrule
\multirow{3}{*}{\rotatebox[origin=c]{90}{Parallel}} &	2	&		55.06 $\pm$ 2.04		&	52.07 $\pm$ 1.60		\\
&	5	&	77.61	$\pm$	1.51	&	70.83	$\pm$	0.96	\\
&	10	&	80.12	$\pm$	0.78	&	70.39	$\pm$	2.28	\\
\bottomrule
\end{tabular}
\end{adjustbox}
\label{tab:multiple_outcomes_fashionmnist}
\end{table}

As described in Section \ref{multiclassdatasets}, we test the robustness of our approach to an increase in number of output classes. We compare \ffl and \textit{parallelized} \ffl across 2-digit, 5-digit and 10-digit classification for \coloredmnist and \coloredfashionmnist. 
\par As shown in Tables \ref{tab:multiple_outcomes_mnist} and \ref{tab:multiple_outcomes_fashionmnist}, both the sequential and the parallel version of \flgames i.e. \ffl and \textit{parallelized} \ffl respectively are robust to an increase in the number of output classes. For both the datasets, \textit{parallelized} \ffl performs at par or better than \ffl.

\begin{table*}[!htb]
\centering
\caption{\small{\coloredfashionmnist, \cifar and \dsprites: Comparison of convergence of methods in terms of mean number of rounds required to reach equilibrium.}}\label{tab:supplement_communication_rounds_results}
\begin{adjustbox}{width=0.9\textwidth}
\begin{tabular}{l|l|l|c|c|c}
\toprule
& & & \multicolumn{3}{c}{Number of Communication Rounds} \\ \cmidrule{4-6}
 & &\sharutalgo & \coloredfashionmnist & \cifar & \dsprites\\
\midrule
\multirow{4}{*}{\rotatebox[origin=c]{90}{Fixed}} & \multirow{2}{*}{\rotatebox[origin=c]{90}{Seq.}} & \ffl & 158.0 $\qquad \;\;\;\;$ & 339.5 $\qquad \;\;\;\;$ & 367.9 $\qquad \;\;\;\;$\\ 
& &\ffictplay & $113.2 \; (1.4 \times)$ & 304.0 $\; (1.1 \times)$ & 257.0 $\; (1.4 \times)$\\ \cmidrule{2-6}
& \multirow{2}{*}{\rotatebox[origin=c]{90}{Par.}} & \ffl & $134.2 \; (1.2 \times)$ & 237.4 $\; (1.4 \times)$ & 233.1 $\; (1.6 \times)$\\
& &\ffictplay & $\textbf{58.4} \;\;\; (2.7 \times)$ & \textbf{217.6} $\; (1.6 \times)$ & \textbf{210.4} $\; (1.8 \times)$\\ 
 \midrule  \midrule 
\multirow{4}{*}{\rotatebox[origin=c]{90}{Variable}} & \multirow{2}{*}{\rotatebox[origin=c]{90}{Seq.}} & \vfl & 454.8 $\qquad \;\;\;\;$ & 557.6 $\qquad \;\;\;\;$ &  628.6$\qquad \;\;\;\;$\\ 
& &\finalalgo & $189.0 \; (2.4 \times)$ & 363.3 $\; (1.5 \times)$ & 455.5 $\; (1.4 \times)$\\ \cmidrule{2-6}
& \multirow{2}{*}{\rotatebox[origin=c]{90}{Par.}} & \vfl & $367.3 \; (1.2 \times)$ & 498.7 $\; (1.1 \times)$ & 520.9 $\; (1.2 \times)$\\
& &\finalalgo & $\textbf{127.8} \; (3.6 \times)$ & \textbf{102.4} $\; (5.4 \times)$ & \textbf{203.0} $\; (3.1 \times)$\\
\bottomrule
\end{tabular}
\end{adjustbox}
\end{table*}


\subsection{Effect of Simultaneous BRD}

\begin{figure}[htb!]
\centering
  \subfloat[]{\includegraphics[width=0.45\columnwidth]{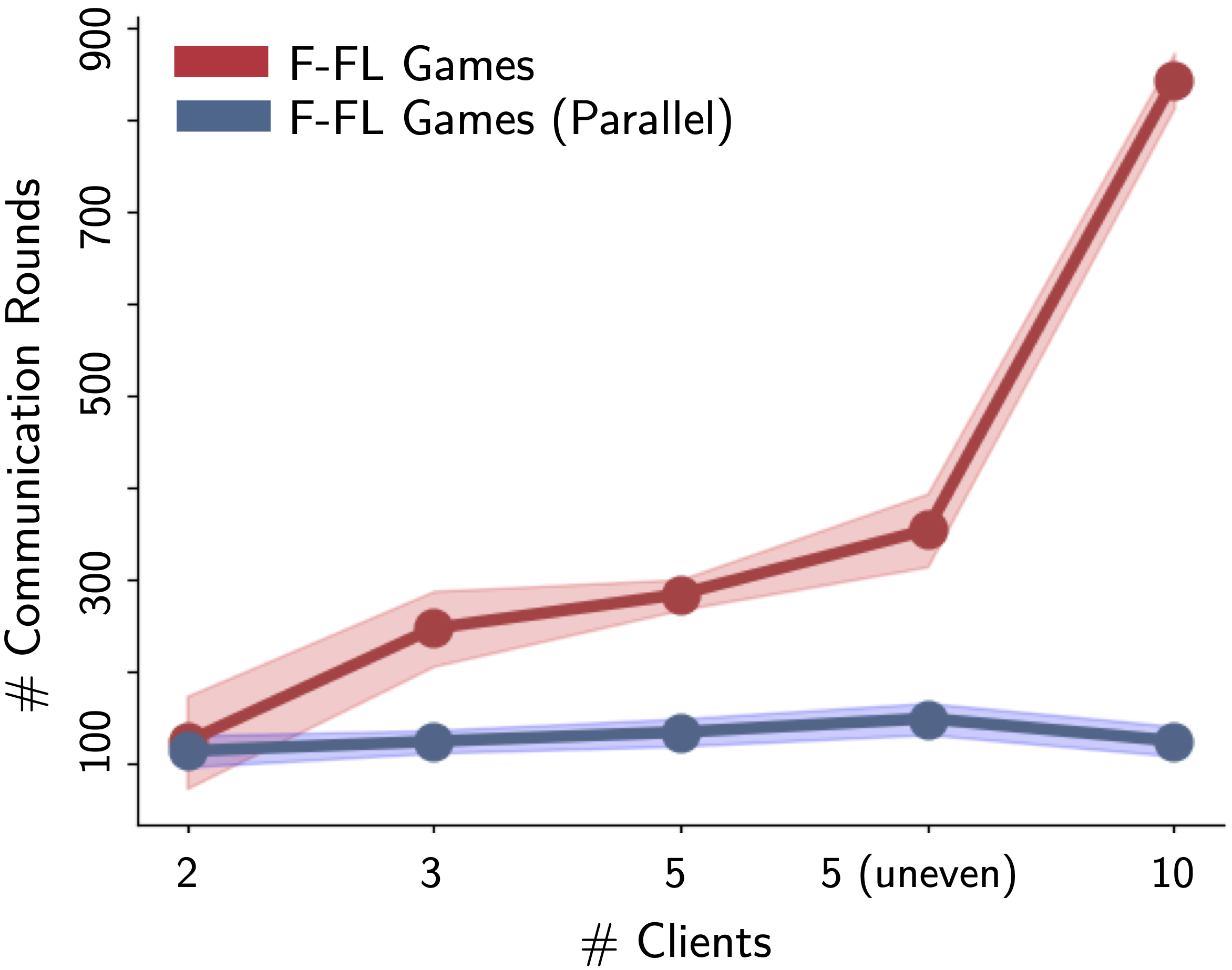}\label{fig:sequential_vs_parallel_ablation1}}
  \hfill
  \subfloat[]{\begin{adjustbox}{width=0.50\columnwidth}
\begin{tabular}{llll}
\toprule
 Type & \# clients & \trainacc & \testacc  \\
\midrule
\multirow{5}{*}{\rotatebox[origin=c]{90}{Sequential}} &	2	&	59.85	$\pm$	7.80	&	65.99	$\pm$	2.44	\\
&	3	&	53.76	$\pm$	5.46	&	67.22	$\pm$	0.93	\\
&	5	&	57.64	$\pm$	2.57	&	67.75	$\pm$	0.52	\\
&	5 (uneven)	&	56.48	$\pm$	2.10	&	67.17	$\pm$	0.84	\\
&	10	&	57.40	$\pm$	1.67	&	68.94	$\pm$	1.35	\\
\midrule
\multirow{5}{*}{\rotatebox[origin=c]{90}{Parallel}}   &	2	&	58.29	$\pm$	4.31	&	69.38	$\pm$	3.61	\\
&	3	&	61.10	$\pm$	1.68	&	69.88	$\pm$	2.82	\\
&	5	&	60.00	$\pm$	5.78	&	68.57	$\pm$	2.86	\\
&	5 (uneven)	&	66.82	$\pm$	2.55	&	66.95	$\pm$	2.41	\\
&	10	&	63.86	$\pm$	3.60	&	70.71	$\pm$	2.99	\\
\bottomrule
\end{tabular}
\end{adjustbox}}
\caption{\small{\coloredfashionmnist: (a) Number of communication rounds required to achieve the Nash equilibrium versus the number of clients in the FL setup; (b) Comparison of \ffl and \ffl (Parallel) with an increase in the number of clients, in terms of training and testing accuracy (mean $\pm$ std deviation).}}
\label{fig:parallel_vs_sequential_fashion_mnist}
\end{figure}

\begin{figure}[htb!]
\centering
  \subfloat[]{\includegraphics[width=0.450\columnwidth]{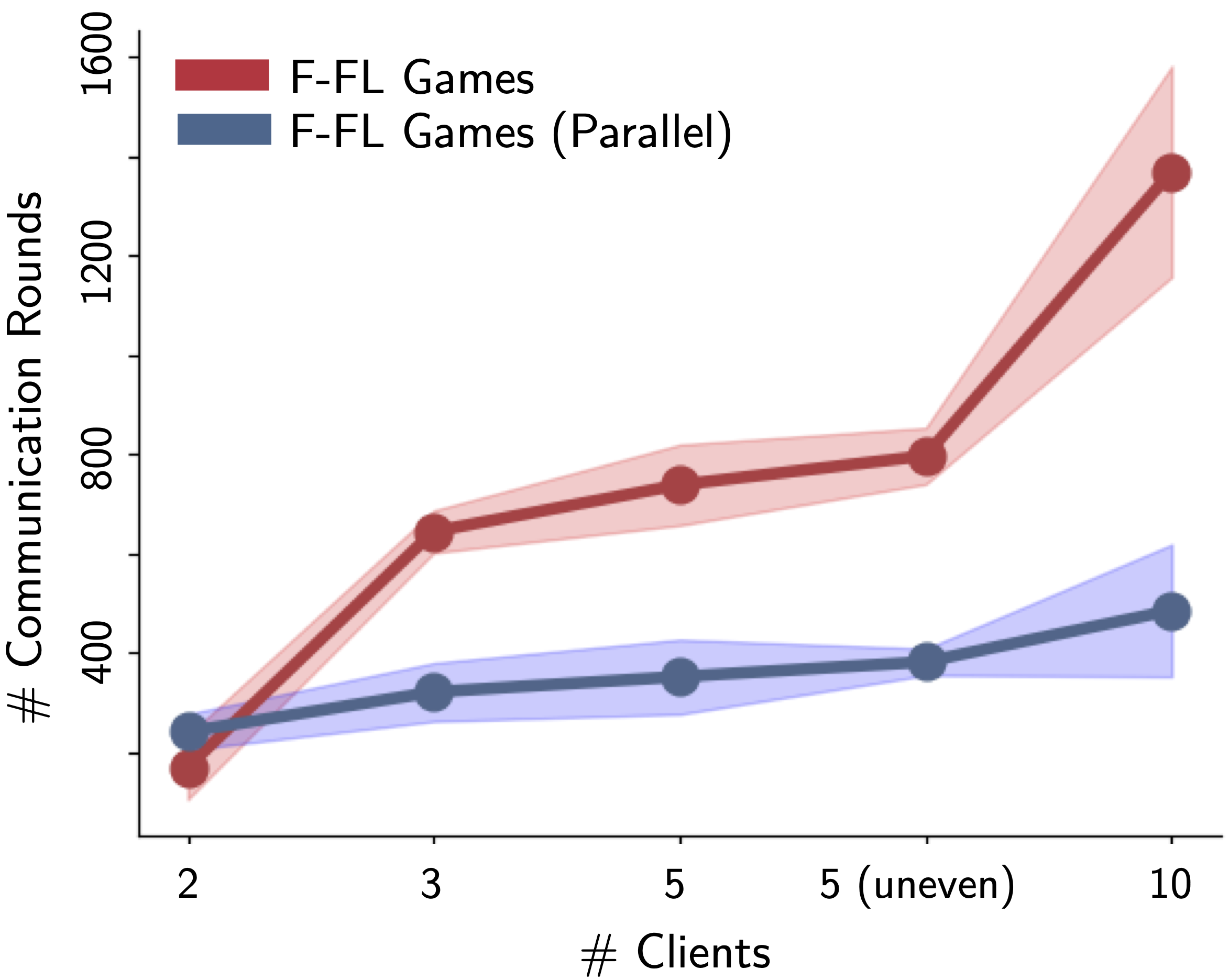}\label{fig:sequential_vs_parallel_ablation2}}
  \hfill
  \subfloat[]{\begin{adjustbox}{width=0.50\columnwidth}
\begin{tabular}{llll}
\toprule
 Type & \# clients & \trainacc & \testacc  \\
\midrule
\multirow{5}{*}{\rotatebox[origin=c]{90}{Sequential}} &	2	&	50.37	$\pm$	3.07	&	48.70	$\pm$	4.12	\\
&	3	&	48.61	$\pm$	1.40	&	59.31	$\pm$	12.45	\\
&	5	&	56.70	$\pm$	6.25	&	45.27	$\pm$	0.42	\\
&	5 (uneven)	&	62.62	$\pm$	6.84	&	47.27	$\pm$	3.81	\\
&	10	&	52.40	$\pm$	1.44	&	50.97	$\pm$	0.92	\\
\midrule
\multirow{5}{*}{\rotatebox[origin=c]{90}{Parallel}}  &	2	&	57.13	$\pm$	4.82	&	50.50	$\pm$	2.70	\\
&	3	&	55.15	$\pm$	2.99	&	53.97	$\pm$	1.06	\\
&	5	&	54.72	$\pm$	1.07	&	51.91	$\pm$	0.50	\\
&	5 (uneven)	&	54.77	$\pm$	2.39	&	53.40	$\pm$	1.83	\\
&	10	&	52.67	$\pm$	1.11	&	53.50	$\pm$	0.84	\\
\bottomrule
\end{tabular}
\end{adjustbox}}
\caption{\small{\cifar: (a) Number of communication rounds required to achieve the Nash equilibrium versus the number of clients in the FL setup; (b) Comparison of \ffl and \ffl (Parallel) with an increase in the number of clients, in terms of training and testing accuracy (mean $\pm$ std deviation).}}
\label{fig:parallel_vs_sequential_cifar}
\end{figure}

\begin{figure}[htb!]
\centering
  \subfloat[]{\includegraphics[width=0.45\columnwidth]{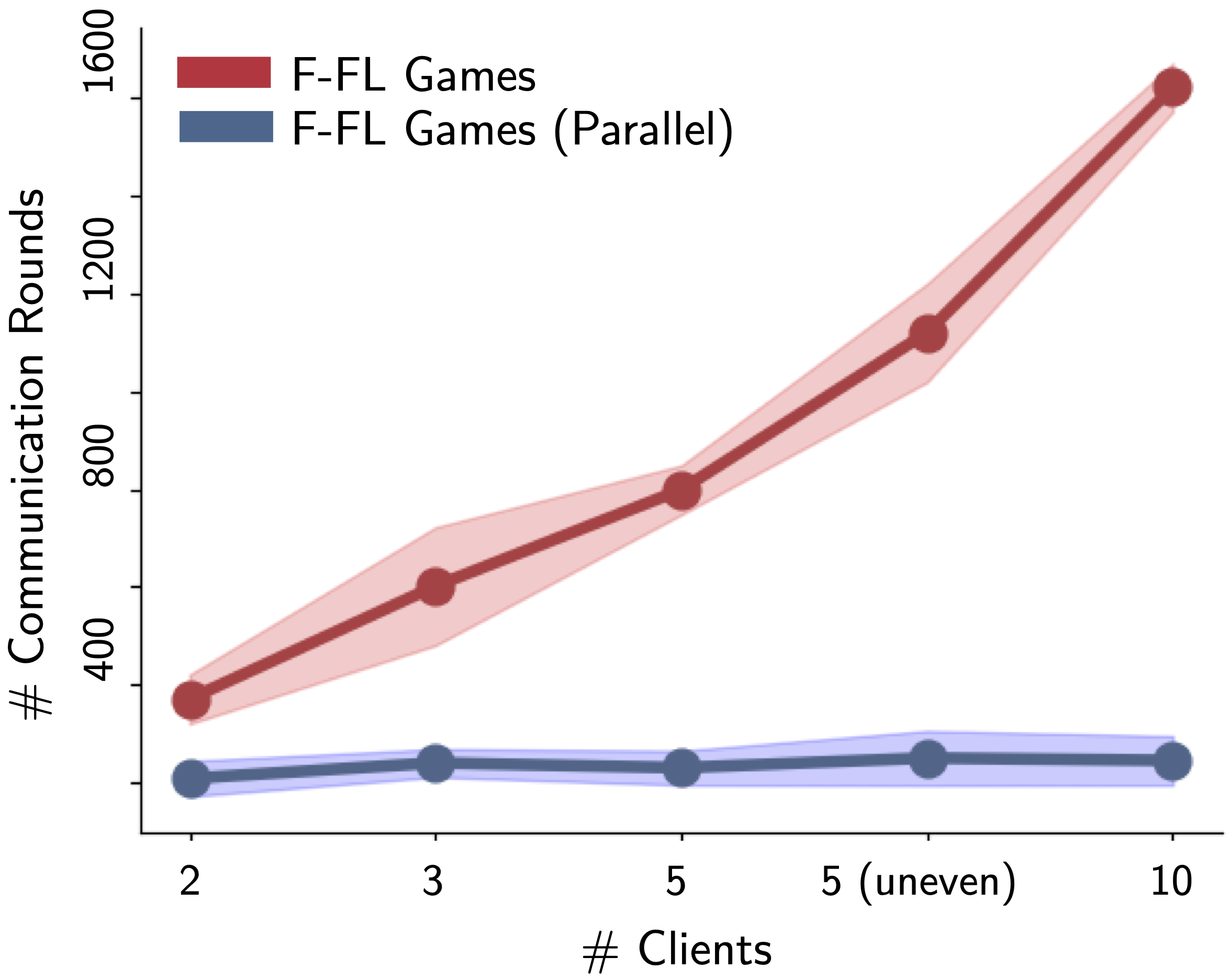}\label{fig:parallel_vs_sequential_lineplot_desprites}}
  \hfill
  \subfloat[]{\begin{adjustbox}{width=.50\columnwidth}
\begin{tabular}{llll}
\toprule
 Type & \# clients & \trainacc & \testacc  \\
\midrule
\multirow{5}{*}{\rotatebox[origin=c]{90}{Sequential}} &	2	&	51.12	$\pm$	4.15	&	54.10	$\pm$	3.45	\\
&	3 & 53.21 	$\pm$ 3.88 & 55.92 	$\pm$ 6.22	\\
&	5& 52.70 	$\pm$ 4.67 & 54.60 	$\pm$ 5.34 \\
&	5 (uneven)	& 54.84 	$\pm$ 2.63 & 55.19 	$\pm$ 3.03\\
&	10	& 53.75 $\pm$ 4.16 & 54.89   $\pm$ 4.20\\
\midrule
\multirow{5}{*}{\rotatebox[origin=c]{90}{Parallel}}  &	2 & 52.31 	$\pm$ 2.78 & 57.30 	$\pm$ 4.02   \\
&	3	& 53.11	$\pm$ 3.69 & 56.04 	$\pm$ 5.20 \\
&	5	& 53.06	$\pm$ 2.94 & 57.01 	$\pm$ 4.23\\
&	5 (uneven) & 52.30	$\pm$ 4.05 & 58.05 	$\pm$ 5.73	\\
&	10	& 53.14	$\pm$ 3.97 & 59.35 	$\pm$ 5.07\\
\bottomrule
\end{tabular}
\end{adjustbox}}
\caption{\small{\dsprites: (a) Number of communication rounds required to achieve the Nash equilibrium versus the number of clients in the FL setup; (b) Comparison of \ffl and \ffl (Parallel) with an increase in the number of clients, in terms of training and testing accuracy (mean $\pm$ std deviation).}}
\label{fig:parallel_vs_sequential_desprites}
\end{figure}

Similar to the experiments conducted for \coloredmnist, where we compared \ffl and \textit{parallelized} \ffl across an increase in the number of clients, we replicate the same setup for \coloredfashionmnist, \cifar and \dsprites. We report the results on the three datasets in Figures \ref{fig:parallel_vs_sequential_fashion_mnist}, \ref{fig:parallel_vs_sequential_cifar} and \ref{fig:parallel_vs_sequential_desprites}. Consistent with the results on \coloredmnist, as the number of clients in the FL system increases, there is a sharp increase in the number of communication rounds required to reach equilibrium. However, the same doesn’t hold true for \textit{parallelized} \ffl. Further, the accuracy achieved by \textit{parallelized} \ffl is comparable or better than that achieved by \ffl. Moreover, as observed from Table \ref{tab:supplement_communication_rounds_results}, across the three benchmarks, all variants of parallel \flgames show significantly faster convergence compared to their corresponding sequential counterparts. This improvement is as large as 3.6$\times$ when comparing \textit{parallelized} \finalalgo and \textit{sequential} \finalalgo.

\subsection{Effect of Memory Ensemble}
\begin{figure}[htb!]
    \centering
\includegraphics[width=0.55\columnwidth]{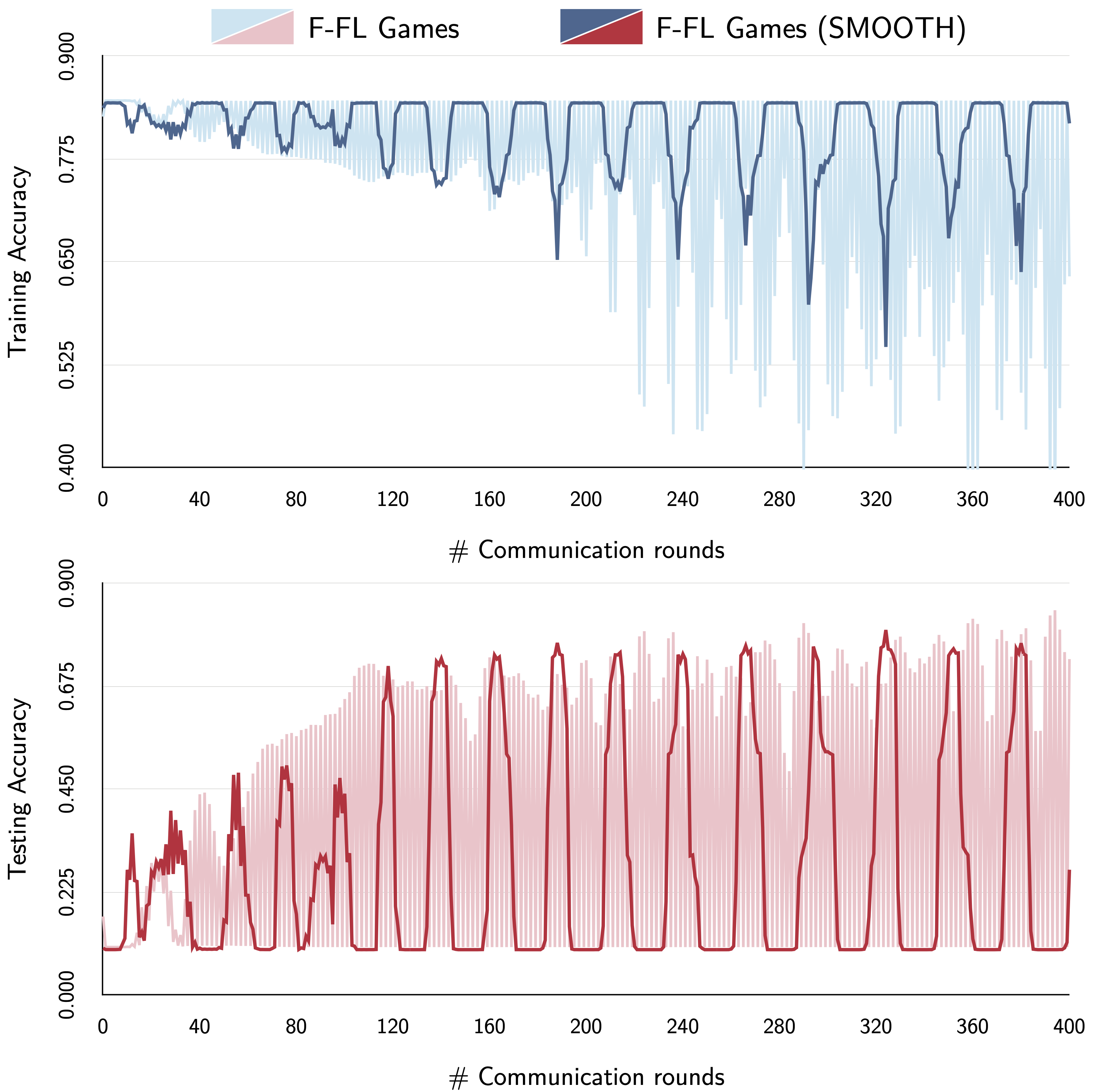}
    \caption{\small{\coloredfashionmnist: Evolution of Training and Testing Training for \ffl and \ffictplay using a buffer size of 5, over the number of communication rounds}}
    \label{fig:fictplay_vs_standard_fashionmnist}
\end{figure}

\begin{figure}[htb!]
    \centering
\includegraphics[width=0.55\columnwidth]{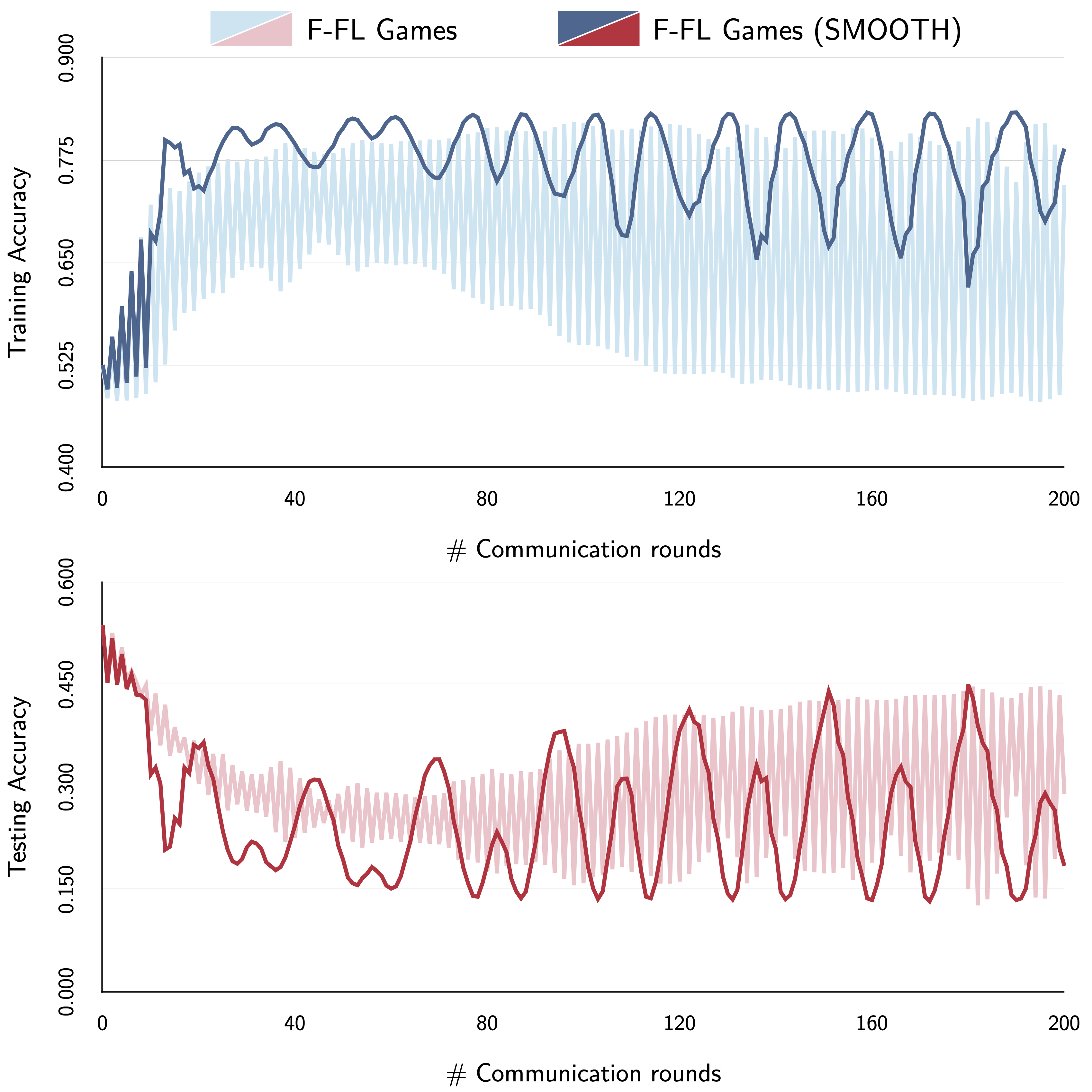}
    \caption{\small{\cifar: Evolution of Training and Testing Training for \ffl and \ffictplay using a buffer size of 5, over the number of communication rounds}}
    \label{fig:fictplay_vs_standard_cifar}
\end{figure}

\begin{figure}[htb!]
    \centering
\includegraphics[width=.55\columnwidth]{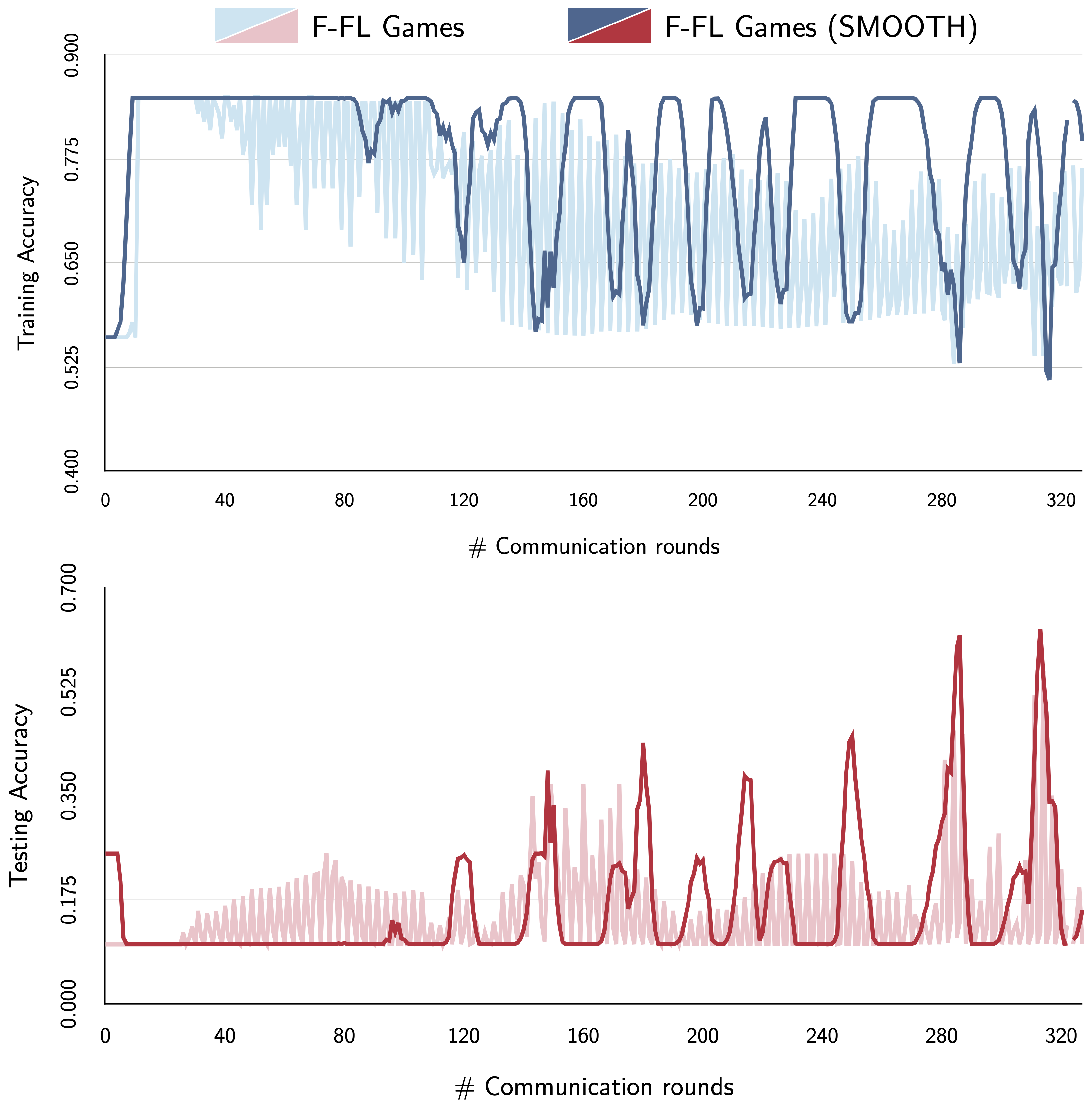}
    \caption{\small{\dsprites: Evolution of Training and Testing Training for \ffl and \ffictplay using a buffer size of 5, over the number of communication rounds}}
    \label{fig:fictplay_vs_standard_desprites}
\end{figure}

Similar to the experiments conducted for \coloredmnist, where we compared \ffl and \ffictplay, we replicate the same setup for \coloredfashionmnist, \cifar and \dsprites. We report the results on all the three datasets in Figures \ref{fig:fictplay_vs_standard_fashionmnist}, \ref{fig:fictplay_vs_standard_cifar} and \ref{fig:fictplay_vs_standard_desprites}.

Consistent with the results on \coloredmnist, performance curves oscillate at each step for \ffl while the oscillations in \ffictplay are observed after an interval of roughly 40 rounds for all the three datasets. Further, \ffictplay also achieves high testing accuracy. This implies that it does not rely on the spurious features to make predictions. As observed from Table \ref{tab:supplement_communication_rounds_results}, apart from achieving high accuracy, both \ffictplay and \textit{parallelized} \ffictplay require significantly fewer communication rounds  as compared to \ffl and \textit{parallelized} \ffl respectively. Moreover, \textit{parallelized} \ffictplay exhibits fastest convergence rate across all the variants of fixed \flgames.
Similar performance evolution curves are also observed for \textit{parallelized} \ffictplay with an added benefit of faster convergence as compared to \ffl and \ffictplay.

\subsection{Effect of using Gradient Descent (GD) for $\phi$}
\begin{figure}[htb!]
    \centering
\includegraphics[width=.55\columnwidth]{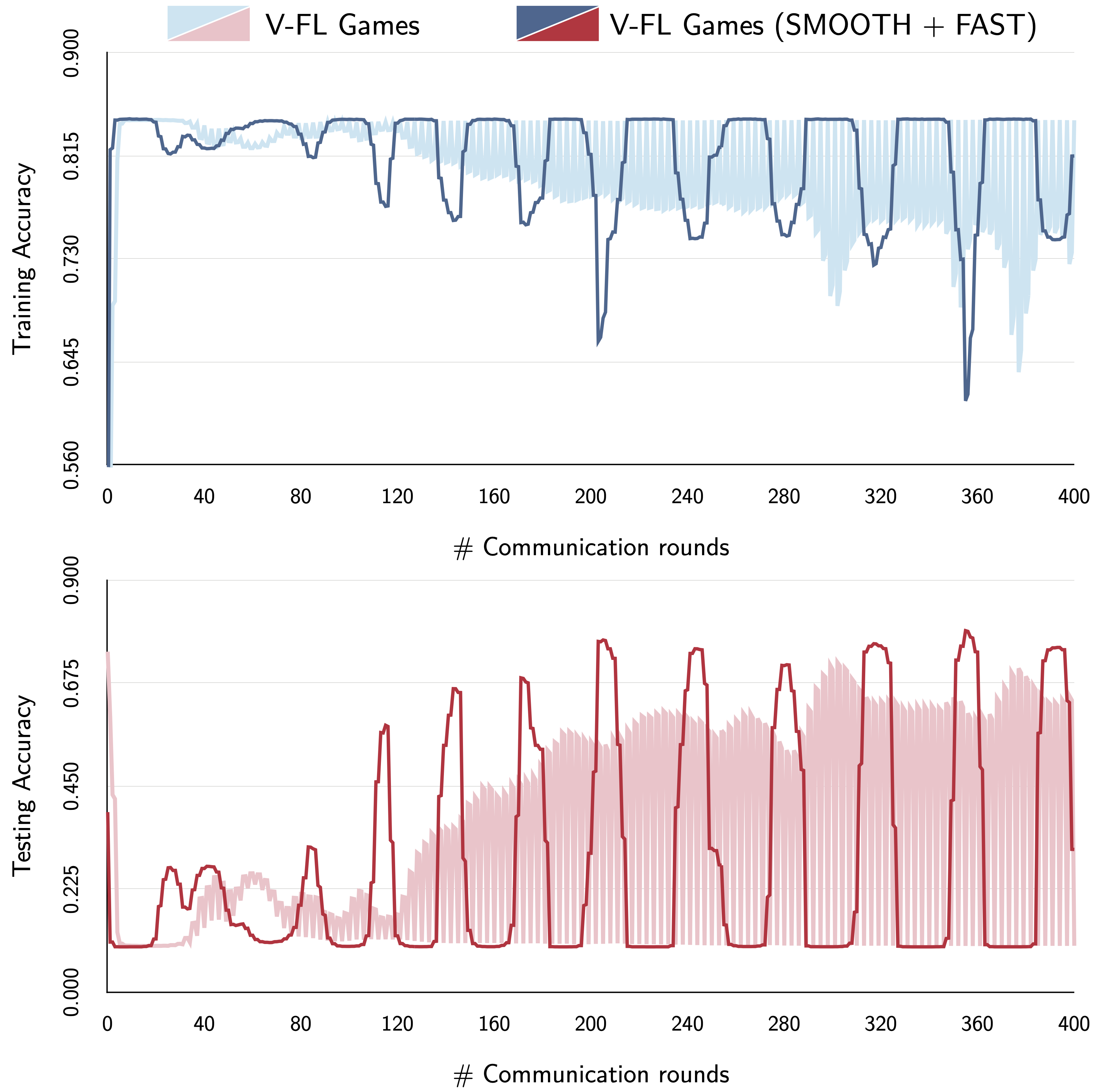}
    \caption{\small{\coloredfashionmnist: Evolution of Training and Testing Training for \vfl and \finalalgo using a buffer size of 5, over the number of communication rounds
    }}
    \label{fig:vfastfictplay_vs_vstandard_fashionmnist}
\end{figure}

\begin{figure}[htb!]
    \centering
\includegraphics[width=.55\columnwidth]{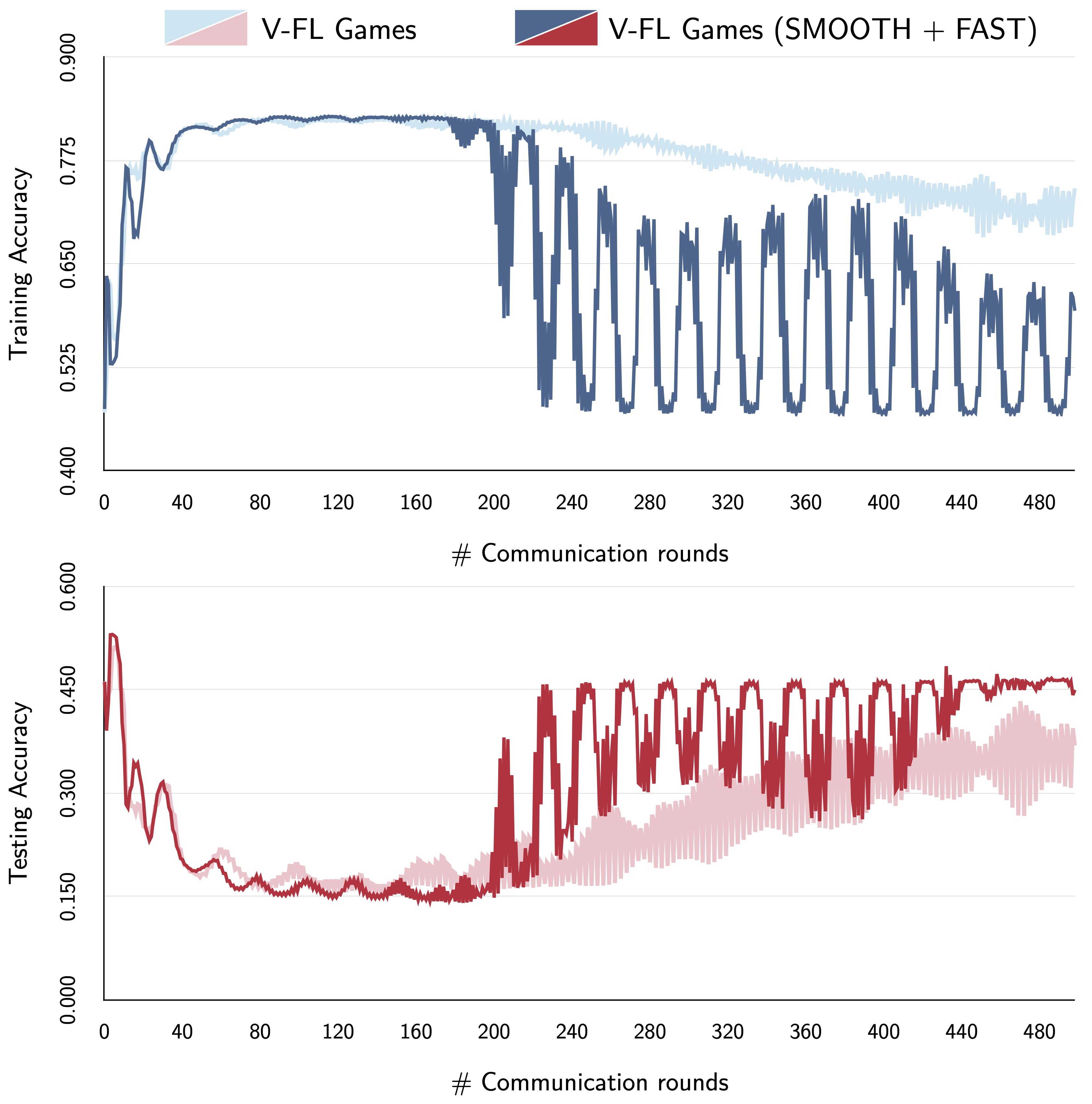}
    \caption{\small{\cifar: Evolution of Training and Testing Training for \vfl and \finalalgo using a buffer size of 5, over the number of communication rounds}}
    \label{fig:vfastfictplay_vs_vstandard_cifar}
\end{figure}

\begin{figure}[htb!]
    \centering
\includegraphics[width=.55\columnwidth]{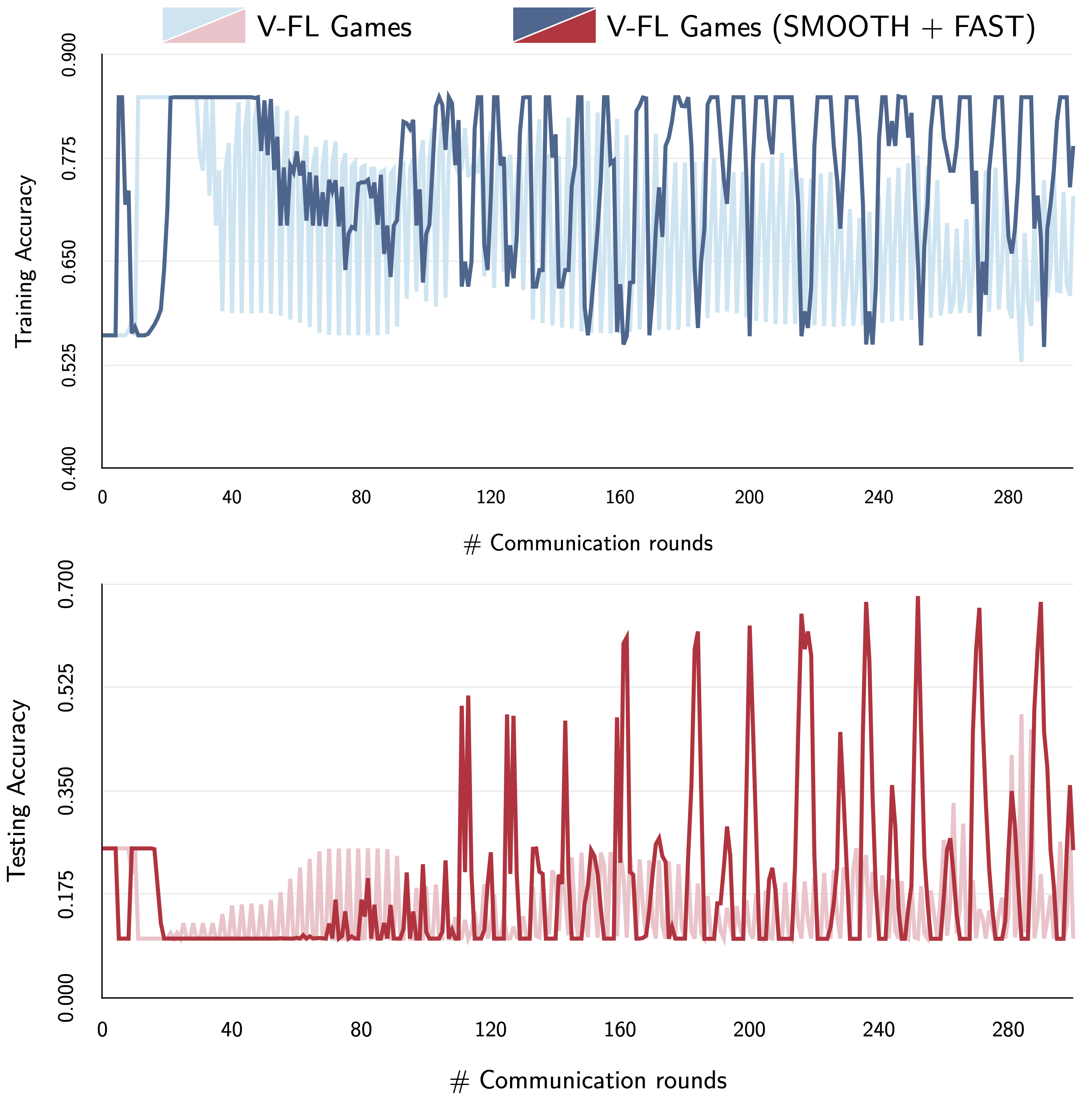}
    \caption{\small{\dsprites: Evolution of Training and Testing Training for \vfl and \finalalgo using a buffer size of 5, over the number of communication rounds}}
    \label{fig:vfastfictplay_vs_vstandard_desprites}
\end{figure}

Similar to the experiments conducted for \coloredmnist, where we compared \vfl and \finalalgo, we replicate the same setup for \coloredfashionmnist, \cifar and \dsprites. We report the results on all the three datasets in Figures \ref{fig:vfastfictplay_vs_vstandard_fashionmnist}, \ref{fig:vfastfictplay_vs_vstandard_cifar} and \ref{fig:vfastfictplay_vs_vstandard_desprites}. Consistent with the results on \coloredmnist, \finalalgo is able to achieve significantly higher testing accuracy in fewer communication rounds as compared to \vfl on all the three benchmarks. Further, as evident from Table \ref{tab:supplement_communication_rounds_results}, \textit{parallelized} \finalalgo has the fastest convergence rate across both \cifar and \dsprites. On a rather simpler dataset i.e. \coloredfashionmnist, \textit{parallelized} \finalalgo exhibits fastest performance across all the variable variants of \flgames. 
\par As discussed before, the performance of variable \flgames is superior to that of it's fixed variant over complex and large-scale datasets like \dsprites. Additionally, as conspicuous from Table \ref{tab:supplement_communication_rounds_results}, \textit{parallelized} \finalalgo requires significantly fewer communication rounds to reach the Nash equilibrium. This further underscores the importance of \textit{parallelized} \vfl both in terms of (i) it's ability to recover the causal mechanisms of the targets, while also providing robustness to distribution shift across clients; (ii) communication efficiency.

\subsection{Effect of exact best response }

\begin{figure}[htb!]
    \centering
\includegraphics[width=.75\columnwidth]{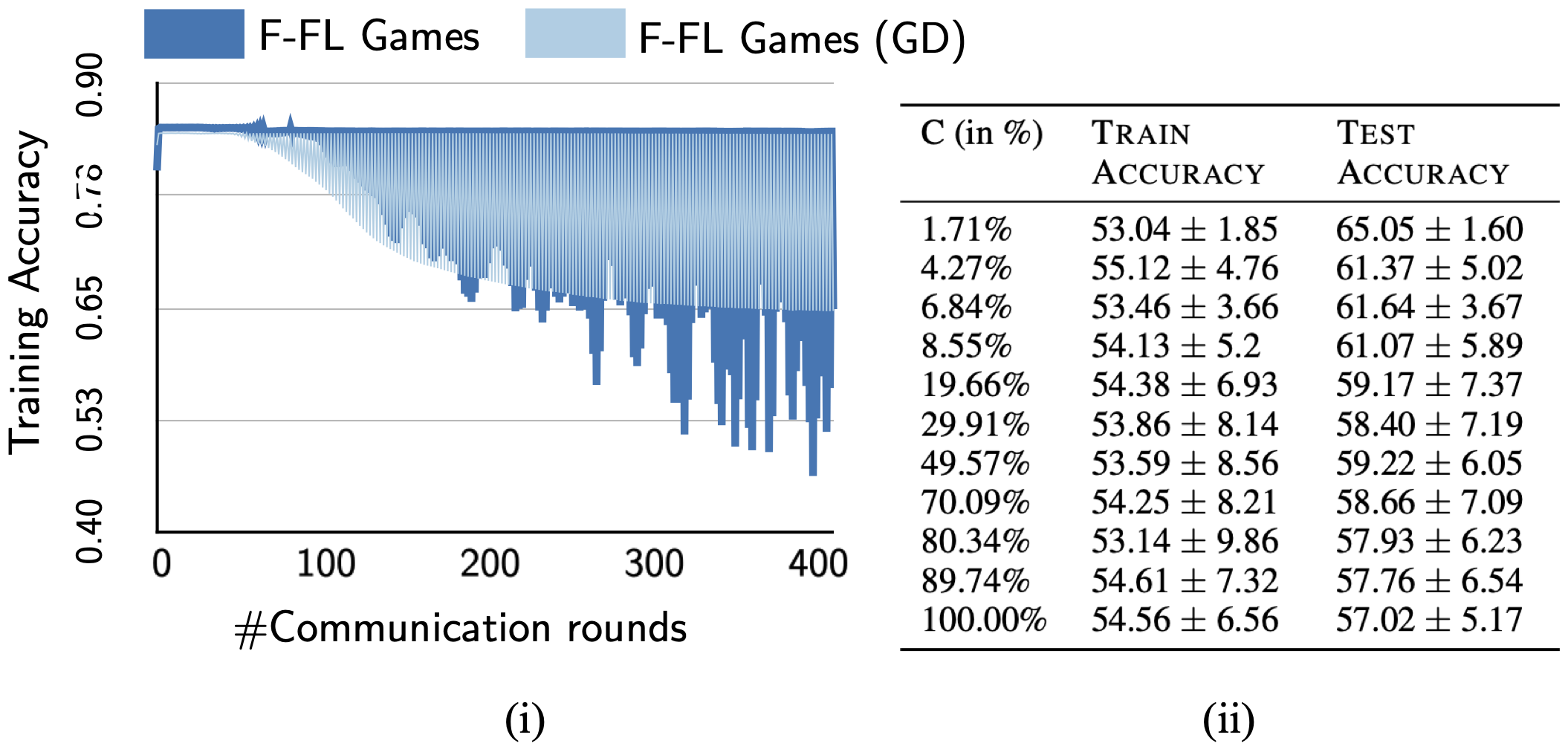}
    \caption{\small{\coloredmnist: (i) Effect on Training accuracy of doing a gradient descent on each client for updating the predictor versus the standard training paradigm i.e. \ffl ;(ii) Impact of increasing the number of local steps (C) for updating the predictor on the training and testing accuracy (mean $\pm$ std deviation).
    }}
    \label{fig:local_compute_vs_standard_mnist}
\end{figure}

\begin{figure}[htb!]
    \centering
\includegraphics[width=.55\columnwidth]{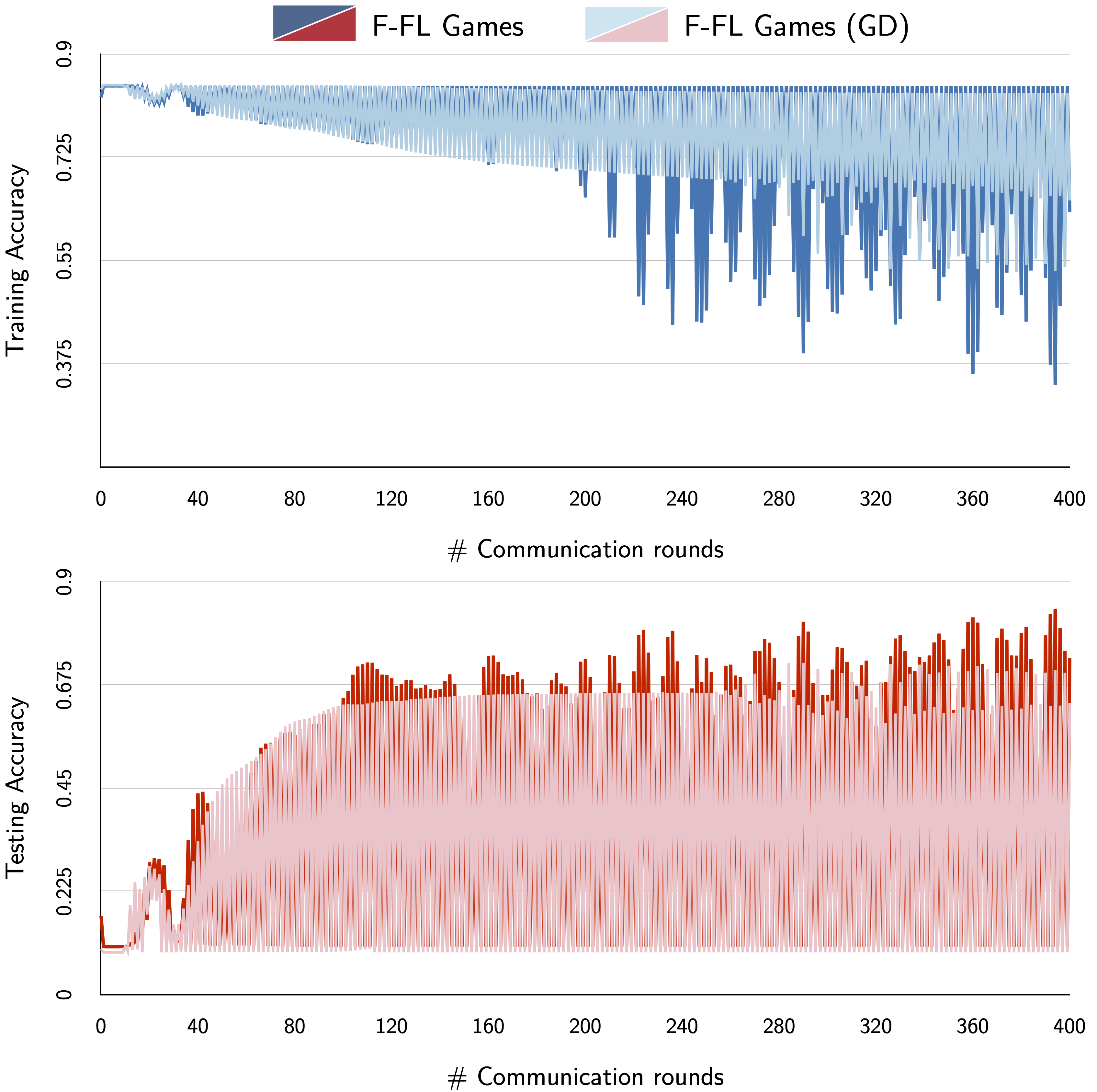}
    \caption{\small{\coloredfashionmnist: Evolution of Training and Testing Training when doing a gradient descent to update the predictor at each client versus the standard training paradigm i.e. \ffl, over the number of communication rounds.
    }}
    \label{fig:local_compute_vs_standard_fashionmnist}
\end{figure}

\begin{figure}[htb!]
    \centering
\includegraphics[width=.55\columnwidth]{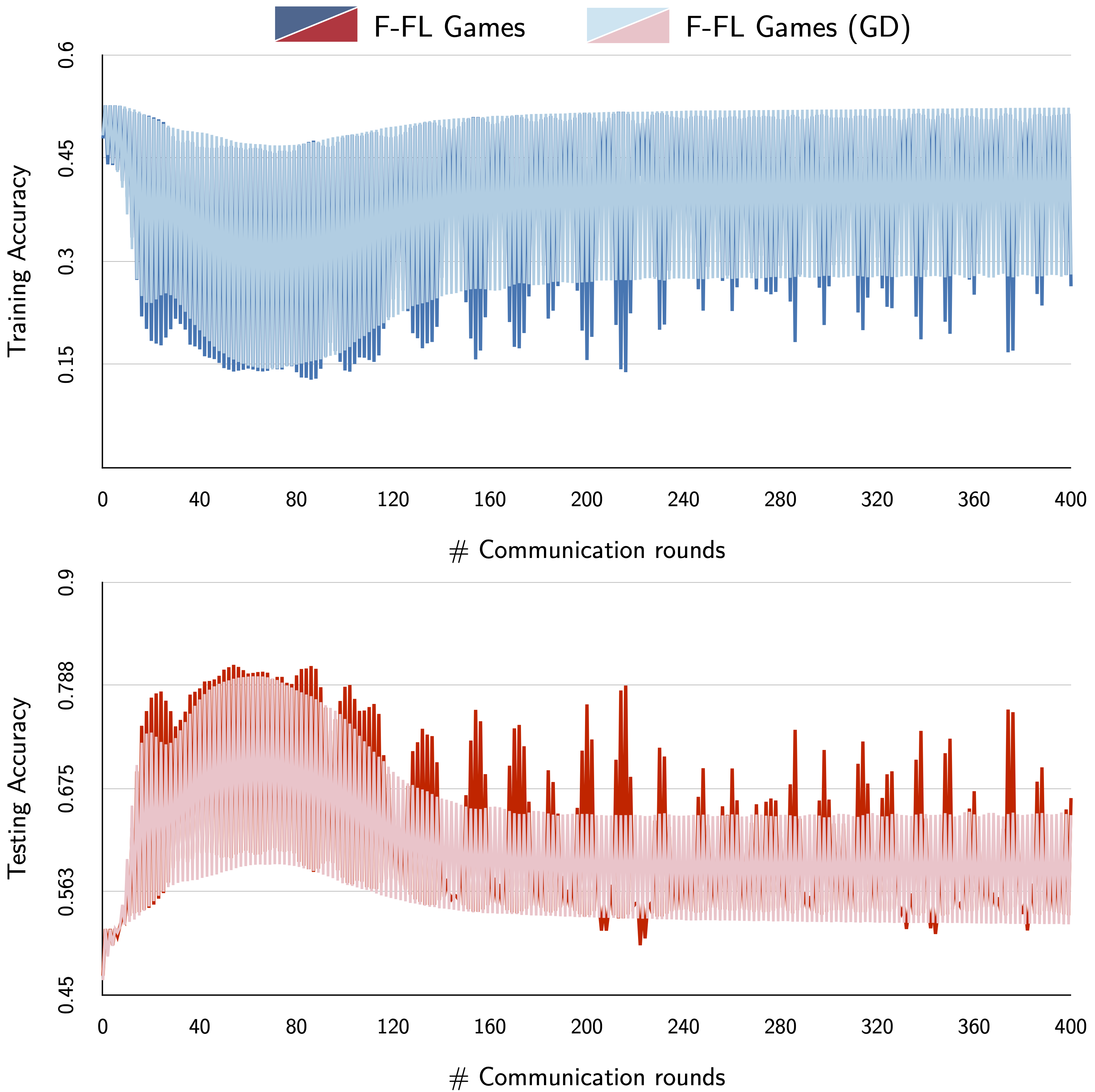}
    \caption{\small{\cifar: Evolution of Training and Testing Training when doing a gradient descent to update the predictor at each client versus the standard training paradigm i.e. \ffl, over the number of communication rounds.}}
    \label{fig:local_compute_vs_standard_cifar}
\end{figure}

\begin{figure}[htb!]
    \centering
\includegraphics[width=.55\columnwidth]{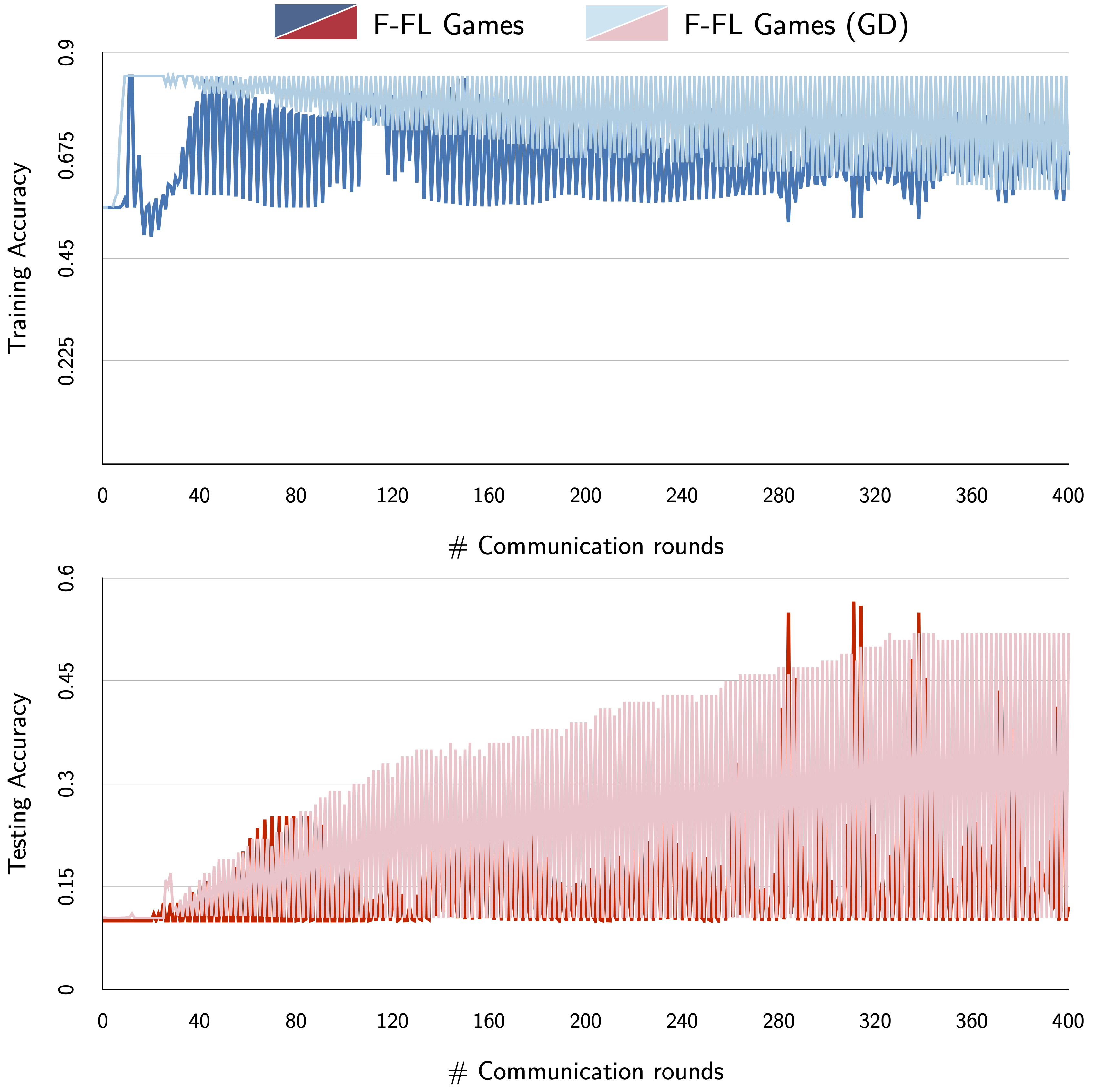}
    \caption{\small{\dsprites: Evolution of Training and Testing Training when doing a gradient descent to update the predictor at each client versus the standard training paradigm i.e. \ffl, over the number of communication rounds.}}
    \label{fig:local_compute_vs_standard_desprites}
\end{figure}
\fedavg provides the flexibility to train communication efficient and high quality models by allowing more local computation at each client. This is particularly detrimental in scenarios with poor network connectivity, wherein communicating at every short time span is infeasible. Inspired by \fedavg, we study the effect of increasing the amount of local computation at each client. Specifically, in \ffl, each client updates its predictor based on a step of stochastic gradient descent over its mini-batch. We modify this setup by allowing each client to run a few steps of stochastic gradient descent locally ($C\%$). When the number of local steps at each client reaches is maximum (training data size/ mini-batch size) or $C=100\%$), the scenario becomes equivalent to a gradient descent (GD) over the training data. For \coloredmnist, it is evident from Figure \ref{local_compute_vs_standard_mnist}(ii) that as the number of local steps increase i.e. each client \textbf{exactly} best responds to its opponents, the testing accuracy at equilibrium starts to decrease. When the local computation reaches 100\%, i.e. each client updates its local predictor based on a GD over its data, \ffl exhibits converges (as shown in Figure \ref{fig:local_compute_vs_standard_mnist}(i)).
\par Similar to the experiments conducted for \coloredmnist, where we studied the influence of increasing the amount of local computation at each client,  we replicate the setup for \coloredfashionmnist, \cifar and \dsprites. Consistent with the results on \coloredmnist and as shown in Figures \ref{fig:local_compute_vs_standard_fashionmnist}, \ref{fig:local_compute_vs_standard_cifar} and \ref{fig:local_compute_vs_standard_desprites}, even when the number of local steps at each client reaches its maximum i.e. (training data size/ mini-batch size)), the trained models are able to achieve high testing accuracy at equilibrium.
Further, this setup achieves faster convergence with higher training accuracy at equilibrium. As observed from the empirical results and discussed by \cite{ahuja2021linear}, \flgames is guaranteed to exhibit convergence and good out-of-distribution generalization behavior \cite{ahuja2021linear} despite increasing local computations. Although the testing accuracy at convergence is lower compared to the standard setup, this approach opens avenues for practical deployment of the approach in FL. 

\end{document}